\documentclass{BAAI_World2026}
\usepackage[utf8]{inputenc} 
\usepackage[T1]{fontenc}    
\usepackage{hyperref}       
\usepackage{url}            
\usepackage{booktabs}       
\usepackage{amsfonts}       
\usepackage{nicefrac}       
\usepackage{microtype}      
\usepackage{marvosym}
\usepackage{mwe}
\usepackage{graphicx}
\usepackage{threeparttable}
\usepackage{amssymb}
\usepackage{adjustbox}
\usepackage{colortbl}
\usepackage{array}
\newcolumntype{P}[1]{>{\centering\arraybackslash}p{#1}}
\usepackage{multirow}
\usepackage{makecell}
\usepackage{bbm}
\usepackage{collcell,xfp}
\usepackage{pgf}
\usepackage{tikz}
\usepackage[most]{tcolorbox}
\usepackage{csquotes}
\usepackage[noorphans,vskip=1em,leftmargin=1em]{quoting}
\usepackage{pifont} 
\usepackage{enumitem} 
\usepackage{tcolorbox} 
\usepackage{forest}
\usepackage{wrapfig}
\usepackage{caption}
\usepackage{subcaption}
\usepackage{longtable}
\usepackage{algpseudocode}
\usepackage{amsmath}
\usepackage[capitalize]{cleveref}
\usepackage[ruled,vlined,linesnumbered]{algorithm2e} 
\usepackage[percent]{overpic}
\usepackage{xspace}
\usepackage[normalem]{ulem}  
\usepackage{tocloft}  
\usepackage[authoryear,round]{natbib}  
\usepackage{tabularx}
\usepackage{listings}
\lstdefinestyle{promptstyle}{
    basicstyle=\rmfamily\footnotesize,
    breaklines=true,
    breakatwhitespace=false,
    columns=fullflexible,
    keepspaces=true,
    frame=single,
    backgroundcolor=\color{gray!5},
    rulecolor=\color{gray!40},
    xleftmargin=1em,
    xrightmargin=1em
}
\newcolumntype{C}[1]{>{\centering\arraybackslash}m{#1}}
\newcolumntype{Y}{>{\raggedright\arraybackslash}X}

\usepackage{fontawesome}

\newcommand{\icona}{\raisebox{-1.5pt}{\includegraphics[height=1.05em]{
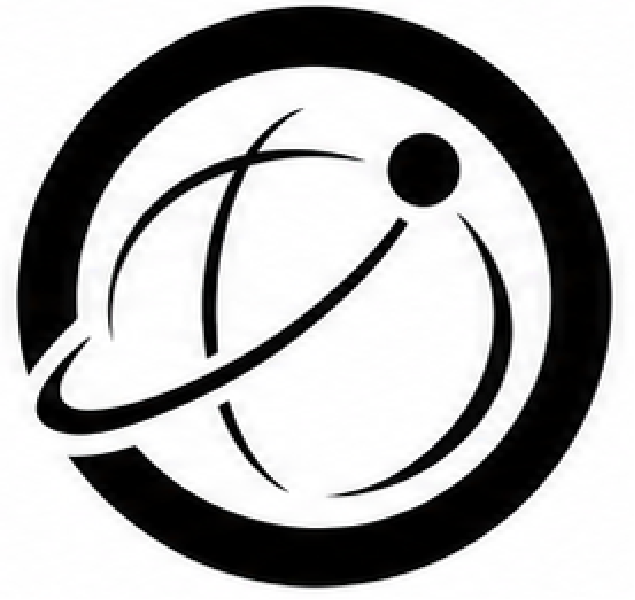}}\xspace}

\definecolor{bluelink}{RGB}{0,75,150}
\hypersetup{
    colorlinks=true,%
    citecolor=bluelink,%
    filecolor=bluelink,%
    linkcolor=bluelink,%
    urlcolor=bluelink
}

\newcommand{\runningtitletext}{}
\usepackage{etoolbox}
\newcommand{\currentnavsection}{sec:intro}
\newcommand{\setnavsection}[1]{%
  \gdef\currentnavsection{#1}%
}

\newcommand{\navseclink}[3]{%
  \begingroup
  \edef\currentsec{\currentnavsection}%
  \def\targetsec{#1}%
  \ifx\currentsec\targetsec
    \hyperref[#1]{\textcolor{black}{\textbf{Sec~#2:~#3}}}%
  \else
    \hyperref[#1]{\textcolor{black}{Sec~#2:~#3}}%
  \fi
  \endgroup
}

\newcommand{\navplainlink}[2]{%
  \begingroup
  \edef\currentsec{\currentnavsection}%
  \def\targetsec{#1}%
  \ifx\currentsec\targetsec
    \hyperref[#1]{\textcolor{black}{\textbf{#2}}}%
  \else
    \hyperref[#1]{\textcolor{black}{#2}}%
  \fi
  \endgroup
}

\newcommand{\sectionheadertext}{%
  \small
  \navseclink{sec:intro}{1}{Intro} |
  \navseclink{sec:method}{2}{Orca} |
  \navseclink{sec:training}{3}{Training} |
  \navseclink{sec:evaluation}{4}{Evaluation} |
  \navseclink{sec:conclusion}{5}{Conclusion} |
  \navseclink{sec:contributors}{6}{Authors} |
  \navplainlink{sec:references}{References}\\
  \navplainlink{sec:appendix_conception}{\textcolor{gray}{Appendix A: Conception}} |
  \navplainlink{sec:appendix_related_work}{\textcolor{gray}{B: Related Work}} |
  \navplainlink{sec:appendix_training}{\textcolor{gray}{C: Train Settings}} |
  \navplainlink{sec:appendix_infrastructure}{\textcolor{gray}{D: Infra}} |
  \navplainlink{sec:appendix_evaluation}{\textcolor{gray}{E: Eval Settings}} |
  \navplainlink{sec:appendix_visualization}{\textcolor{gray}{F: Visualization}}
}

\fancypagestyle{mainstyle}{
    \fancyhf{}
    \fancyhead[C]{%
        \vbox{%
            \centering
            {\normalfont\bfseries\fontsize{9.5}{11}\selectfont \runningtitletext\par}%
            \vspace{1pt}
            {\normalfont\fontsize{9.5}{11}\selectfont \sectionheadertext\par}%
            \vspace{3pt}
            {\color{black}\hrule height 0.5pt}%
        }%
    }

    \fancyfoot[C]{\footerfont \thepage}

}

\usepackage{titlesec}

\definecolor{navyblue}{HTML}{0071BC}
\iftrue   
\newcommand{\displaytodo}[1]{#1}
\else
\newcommand{\displaytodo}[1]{}
\fi

\definecolor{blindcolor}{HTML}{AB2AC6}    
\definecolor{chancecolor}{HTML}{F59E0B}   
\definecolor{singlecolor}{HTML}{06B6D4}   
\definecolor{multiplecolor}{HTML}{2563EB} 
\definecolor{captioncolor}{HTML}{22C55E}  

 \newcommand{\culine}[2]{%
    \def\temp@uline{\bgroup\markoverwith
        {\textcolor{#1}{\rule[-0.5ex]{2pt}{1pt}}}\ULon}%
    \temp@uline{#2}%
}
 \newcommand{\cthickuline}[3][0.8pt]{%
    \def\temp@uline{\bgroup\markoverwith
        {\textcolor{#2}{\rule[-0.5ex]{2pt}{#1}}}\ULon}%
    \temp@uline{#3}%
}

\title{\center{Orca: The World is in Your Mind}}

 \renewcommand\Affilfont{\normalfont\fontsize{11}{11}\selectfont} 
\author{Orca Team, Beijing Academy of Artificial Intelligence}

\newcommand{\linktable}{
\begin{center}
    \renewcommand{\arraystretch}{1}
    \begin{tabular}{@{}r@{\hspace{0.5em}}l@{\hspace{1.2em}}l@{}}
        \icona{} & \textbf{Website} & \url{https://orca-wm.github.io}\\
    \end{tabular}
\end{center}
}

\begin{abstract}
We introduce Orca, an initial instantiation of a general world foundation model. Orca learns a unified world latent space from multimodal world signals and exposes it through multimodal readout interfaces.
Rather than optimizing isolated next-token, next-frame, or next-action prediction, we are centered on \textit{Next-State-Prediction} modeling, offering a unified state-transition modeling route toward understanding, predicting, and acting upon the world.
Orca learns through two complementary paradigms: \textit{unconscious learning} captures dense natural state transitions from continuous videos, and \textit{conscious learning} models sparse meaningful state transitions by language-described events and VQA supervision. 
For pre-training, we construct a large-scale world-learning inventory data, including 125K hours of video data and 160M event annotations. 
After pre-training, Orca learns a unified world latent space. 
To examine whether the learned latent supports downstream, we evaluate it by three representative downstream readouts: text generation, image prediction, and embodied action generation. Orca's backbone is frozen, and only the lightweight modality-specific decoders are trainable.
Experiments show the scalability of the proposed paradigm and verify that stronger world latent enables stronger downstream readouts. Orca outperforms similar-sized specialized baselines.
These results show that Orca, as a general world foundation model, presents a promising approach to understanding, predicting, and acting upon the world. 
Finally, we discuss the current limitations, aiming to provide useful insights and inspiration for the community.
\end{abstract}

\newcommand{\abstractboxwidth}{0.9\linewidth}
\makeatletter
\renewcommand{\abscontent}{%
    \begin{center}
        {\color{black}\fontsize{15pt}{14pt}\selectfont\textbf{Abstract}\par}%
        \vspace{2ex}%
        \parbox{\abstractboxwidth}{\absfont \theabstract}%
        \@ifundefined{@keywords}{}{%
            \vskip1em
            \parbox{\abstractboxwidth}{\keywordsfont Keywords: \@keywords}%
        }%
    \end{center}%
}
\makeatother

\begin{document}
\pagestyle{mainstyle}
\let\storedabscontent\abscontent

\maketitle

\vspace{2ex}
\begin{figure}[!htbp]
    \centering 
    \includegraphics[width=1\linewidth]{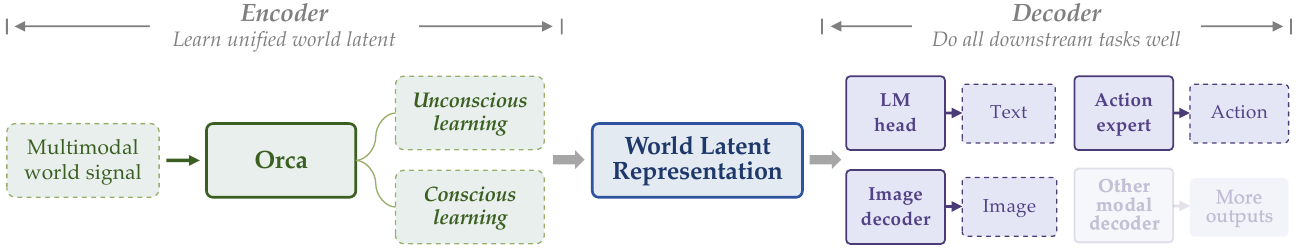}
    \caption{\textbf{The Orca's overall framework.} Orca follows an \textit{\textbf{Encoder-Decoder}} architecture. Given multimodal world signals, the \textit{\textbf{Encoder}} learns a world latent through two complementary paradigms: \textit{unconscious learning} and \textit{conscious learning}. Unconscious learning captures dense natural state transitions, while conscious learning captures sparse meaningful state transitions. To prove that the learned latent is effective, the \textbf{\textit{Encoder}} is frozen after pre-training, and only the lightweight modality-specific decoders are trainable separately. The \textit{\textbf{Decoder}} reads out the latent into text, images, actions, and other modalities.}
\label{fig:teaser}
\end{figure}



\section{Introduction}\label{sec:intro}
\setnavsection{sec:intro}

We argue that the essential next step toward general intelligence is to build a model that can continuously learn and self-evolve like a human, and ultimately transcend human cognitive boundaries. As it internalizes physical laws, causal relationships, and dynamic evolution, this model is expected to develop into a self-emerging intelligence system. Such a model should continuously absorb multimodal world signals to model the world's latent states. Ideally, these signals should encompass \textit{neural signals} such as vision, text, audio, action, and tactile; also include \textit{physical signals} such as force and light; and even include \textit{signals from fields} such as the macrocosm, the microscopic quantum systems, and life sciences.
More importantly, such a model should use state-transition modeling as a unified paradigm for both observed and unknown domains, thereby opening new possibilities for exploring the world.

From this perspective, intelligence should not merely be \textit{Next-Token-Prediction} model that can respond to instructions \citep{DeepSeek-V4, qwen36_35b_a3b, Emu3, GPT54}, \textit{Next-Frame-Prediction} model that can generate high-quality images and videos \citep{nanobananapro, ChatGPT-Images-2.0, seedance2.0}, or \textit{Next-Action-Prediction} model that can generate high-quality action \citep{pi0.7, DreamZero, GR00TN1.7}. Instead, it should be defined by the ability to build world states and support the latent space for diverse downstream tasks. These points toward a general world foundation model grounded in \textit{\textbf{Next-State-Prediction}} modeling, and including \textit{implicit dynamics} and \textit{explicit conditions}.
Our conception visualization is shown in \textbf{Appendix \ref{sec:appendix_conception}}.

We present Orca, a world learner that takes an initial step toward the above goal by learning a world latent space. Figure \ref{fig:teaser} shows the Orca's overall framework. In this version, \textbf{\textit{Encoder}} focuses on two fundamental signal types: visual and language. Visual signals, including videos and images, are similar to how humans perceive the world. Language signals correspond to how humans understand the world, providing causal explanations and task intentions. Orca has two learning paradigms to realize:

\vspace{-1ex}
\begin{enumerate}[label={}, leftmargin=10pt, itemindent=0pt, itemsep=0.3em, topsep=0.3em]
    \item[] \textit{\textbf{1) Unconscious learning} aims to learn natural and dense state transitions from continuous video.} \ This process does not rely on labeled tags, but instead uses the supervision provided by itself. The model learns natural evolution by predicting the latents of the next frames and internalizes state transitions. 
    \item[] \textit{\textbf{2) Conscious learning} aims to learn meaningful and sparse state transitions under the constraints of instructions.} \ The model uses textual constraints to learn meaningful state transitions at the event level.
\end{enumerate}

Orca builds a world latent space through the two paradigms. The \textbf{\textit{Decoder}} reads out text, images, and actions. \textbf{\textit{Note}} that these readouts are not intended to chase task-specific SOTA performance, but to examine two core questions: \textit{1) the proposed paradigms are feasible and scalable}, and \textit{2) stronger world modeling leads to stronger downstream readouts.} Therefore, the Orca backbone is frozen during decoder post-training, and only lightweight readout modules are trainable. Experiments answer these questions: Orca's training losses continue to decrease with model size and data scale, and its language, image, and action readouts consistently improve as pre-training scales up. The main contributions are as follows:

\vspace{-0.5em}
\begin{itemize}
    \item \textit{We propose Orca.} \ Orca learns a world latent space from multimodal world signals. This latent space can serve as a general interface for multimodal downstream readouts.

    \item \textit{We design two complementary learning paradigms.} \ Unconscious learning captures natural dense state transitions from continuous videos, while conscious learning leverages textual conditions to learn meaningful sparse state transitions associated with decisions and task outcomes.
    
    \item \textit{We construct a large-scale collection to support Orca's learning paradigm.} \ We built the inventory data that contains 125K hours of videos and 160M event annotations, covering ego-centric interaction, exo-centric manipulation, action-free robot execution, and event-level transitions for pre-training.
    
    \item \textit{Experiments show that Orca learns the effective world latent.} \ Experiments demonstrate the scalability of the proposed paradigm and show through readout probing that a stronger world latent enables stronger downstream capabilities. Across text generation, real-world interactive image prediction, and embodied action generation, Orca outperforms specialized baselines at a comparable scale. We further provide a careful analysis of its current limitations, aiming to offer insights and inspiration for the community's future sustainable development.
\end{itemize}


\section{Orca}\label{sec:method}
\setnavsection{sec:method}

\subsection{Modeling}
\label{sec:method_modeling}

\paragraph{Macro.} Orca formulates world learning as latent world-state modeling, including state abstraction from multimodal world signals and state transition.
Given the world signals $\mathcal{X}, \mathcal{X}=\{X^{m}\}_{m\in\mathcal{M}}$, where $\mathcal{M}$ can encompass a rich variety of modalities. Ideally, $\mathcal{X}$ should include all signals present in the world, such as \textit{common multimodal signals}: language, vision, and audio; \textit{physical signals}: force and light; and even \textit{signals beyond human perception}: infrared radiation. Mapping $\mathcal{X}$ to a latent world state $\mathcal{S}$, i.e., $\mathcal{S} = f_{\theta}(\mathcal{X}).$
We model the state $S\in \mathcal{S}$ evolves forwards and backwards under \textit{implicit dynamics} and \textit{explicit conditions}:
\begin{equation}
\label{eq:Orca_state_transition}
    S_{t+\Delta}\sim p_{\Theta}
    \left(S_{t+\Delta}\mid S_t,z_t,c_t\right),
    \quad \Delta\in\mathbb{Z}_{\ne 0}.
\end{equation}
where, $z_t$ is a way to realize \textit{invisible dynamics}. It captures latent or unobserved factors that drive state changes, such as physical laws, object properties, scene dynamics, and environmental forces. $c_t$ is a way to realize \textit{explicit conditions}. It refers to observed conditions such as human instructions. $\Delta>0$ represents predicting future states $S_{> t}$, while $\Delta<0$ represents backtracking to past states $S_{< t}$.

\paragraph{Details.} 
In this version, Orca uses visual signals and language signals as two fundamental types of multimodal world signals.
To realize the state-transition modeling in Equation~\ref{eq:Orca_state_transition}, Orca adopts two complementary learning paradigms: \textit{unconscious learning} and \textit{conscious learning}.

\vspace{-1ex}
\begin{enumerate}[label={}, leftmargin=10pt, itemindent=0pt, itemsep=0.3em, topsep=0.3em]

    \item[] \textit{\textbf{1) Unconscious learning} learns state transitions from observation alone.} \ The model observes naturally occurring transitions without explicit semantic conditions. It can learn how objects move, natural dynamics, physical regularities, or how scenes transition over time. This is equivalent to $c_t=\varnothing$, i.e., $S_{t+\Delta}\sim p_{\Theta}^{\mathrm{u}}\left(S_{t+\Delta}\mid S_t,z_t\right)$. In this paradigm, the target state originates from the nearest future observation.
    \textit{Unconscious learning} can learn dense and natural state transitions.

    \item[] \textit{\textbf{2) Conscious learning} learns state transitions under explicit semantic conditions.} \ Given a language signal, Orca treats it as an explicit condition that guides the state transition. The language condition can specify an event $c_t=e_{t+\Delta}$, including a future or past event, i.e., $S_{t+\Delta}\sim p_{\Theta}^{\mathrm{c}}(S_{t+\Delta}\mid S_t, z_t,e_{t+\Delta})$. The condition can also be a task intention or a causal premise. It guides how the current state should transition toward a target state. \textit{Conscious learning} can learn sparse and meaningful state transitions.

\end{enumerate}

\subsection{Architecture}
\label{sec:method_architecture}

\begin{figure}[!htbp]
    \centering
    \includegraphics[width=1\linewidth]{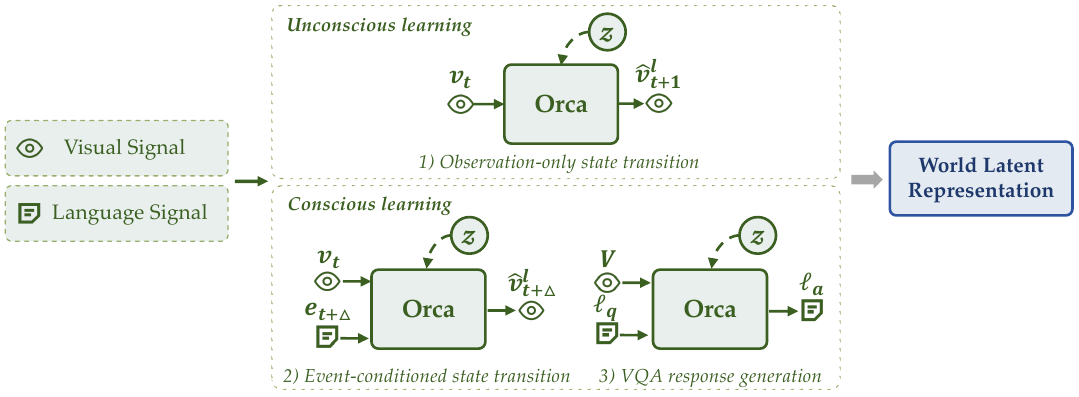}
    \caption{\textbf{Overview of Encoder.} Orca learns a world latent representation through two learning paradigms. \textit{Unconscious learning} uses video data to capture dense and natural state transitions. \textit{Conscious learning} uses language instructions as explicit semantic conditions to capture sparse and meaningful state transitions.}\label{fig:Orca_architecture}
    \vspace{-0.5em}
\end{figure} 

\paragraph{Encoder.} 
Orca focuses on the \textit{\textbf{Encoder}}, which learns a unified world latent space for state abstraction and state transition.
The overview of the \textit{\textbf{Encoder}} is shown in Figure \ref{fig:Orca_architecture}. It uses a native pre-trained VLM \citep{qwen35} aligned with the language and vision spaces. Given visual and language signals, Orca learns the world latent through \textit{unconscious learning} and \textit{conscious learning}.

The input of \textit{unconscious learning} is a certain frame $v_t$ of the video $V$, and the output is the prediction latent $\hat{v}^l_{t+1}$ of the next adjacent frame. After passing through the VLM, $v_t$ will be used to obtain the predicted $\hat{v}^l_{t+1}$ through two layers of MLP. The ground truth of the next adjacent frame $v_{t+1}$ will only pass through the frozen vision encoder to obtain a latent representation $v^l_{t+1}$, and then be teacher forced with the prediction latent $\hat{v}^l_{t+1}$. This part completes the \textit{1) Observation-only state transition} in Figure \ref{fig:Orca_architecture}.

To support \textit{conscious learning}, we divide the video into multiple segments based on meaningful events. Each event contains a video segment and a corresponding instruction description, as shown in \textbf{Section \ref{sec:pretraining_data}}. The input of \textit{conscious learning} also includes a frame $v_t$ from $V$, along with an instruction description $e_{t+\Delta}$ of the adjacent (next or previous) event, where $v_t$ belongs to a certain event. The output is the prediction latent $\hat{v}^l_{t+\Delta}$ of a random sample associated with the event $e_{t+\Delta}$. This part completes the \textit{2) Event-conditioned state transition} in Figure \ref{fig:Orca_architecture}. In addition, the essence of conscious learning is understanding the world. Therefore, we will also input the video $V$ and the related questions $l_q$, and output the corresponding language answers $l_a$. This part completes the \textit{3) VQA response generation} in Figure \ref{fig:Orca_architecture}. Ultimately, the world latent space is obtained by the two learning paradigms. 

\paragraph{Decoder.} The learned latent space is read out by the modality-specific \textbf{\textit{Decoder}} to extract multi-modal information. Since the decoder is not the focus of this section, its details will be shown in \textbf{Section \ref{sec:downstream_readout}.}


\section{Training}\label{sec:training}
\setnavsection{sec:training}

Orca is trained in two stages. The \textbf{\textit{pre-training}} stage learns the world latent through large-scale visual and language data. In the \textbf{\textit{downstream post-training}} stage, the Orca's backbone is frozen. Only modality-specific readout modules are trainable to obtain language, vision, and action information. 

\subsection{Pre-Training}\label{sec:pretraining}

\subsubsection{Pre-Training Recipe}
\label{sec:pretraining_recipe}

Orca pre-training instantiates world-state modeling with three objectives: \textit{1) observation-only state transition}, \textit{2) event-conditioned state transition}, and \textit{3) VQA response generation}. The two state-transition objectives are implemented through a set of learnable query vectors, while \textit{3) VQA response generation} is optimized through the language modeling (LM) head of the backbone.

\paragraph{Learning Objectives.}
We instantiate pre-training with the three objectives. \textit{1) observation-only state transition} and \textit{2) event-conditioned state transition} are implemented with learnable queries in the input of the VLM backbone. The input is: \textit{<visual token>, <Query 1>, <Instruction>, <Query 2>}. \textit{Note that all learnable queries are trained from scratch.} The specific implementations are shown in \textbf{Appendix~\ref{sec:appendix_pretraining_query_implementation}}.

\vspace{-1ex}
\begin{enumerate}[leftmargin=1.35em, itemsep=0.35em, topsep=0.35em, parsep=0em]
    \item[] \textbf{\textit{1) Observation-only state transition.}} This objective forms \textit{unconscious learning}. Given \(v_t\), Orca takes \(v_t\) together with the \textit{<Query 1>} \(q_1\). The last-layer hidden state of \(q_1\) is passed through two layers of MLP to predict $\hat{v}^l_{t+1}$. The ground truth $v^l_{t+1}$ is obtained through the frozen vision encoder. Continuous videos provide dense supervision, allowing the model to capture naturally occurring world dynamics such as motion, occlusion, object interaction, and scene changes.
  
    \item[] \textbf{\textit{2) Event-conditioned state transition.}} This objective forms \textit{conscious learning} with language. Given \(v_t\), \textit{<Query 1>} \(q_1\), $e_{t+\Delta}$, and \textit{<Query 2>} \(q_2\), the last-layer hidden state of \(q_2\) is passed through the two layers of MLP to predict $\hat{v}^l_{t+\Delta}$. The ground truth $v^l_{t+\Delta}$ is also obtained through the frozen vision encoder. $e_{t+\Delta}$ describes an event, a task intention, or a causal premise. 

    \item[] \textbf{\textit{3) VQA response generation.}} This objective provides another path for common sense in \textit{conscious learning}. Given $V$ and $l_q$, Orca produces $l_a$ with the standard next-token prediction loss.
\end{enumerate}

The first two objectives are supervised in the latent space of the vision encoder. This design focuses on pre-training for state modeling rather than pixel-level reconstruction. The last objective uses the LM head. Given $V$ and $l_q$, the LM head predicts $l_a$ with the standard next-token prediction loss.

\paragraph{Training Components.}

The full pre-training loss combines the three components as:
\begin{equation}\label{eq:total_loss}
\mathcal{L}=\lambda_{\mathrm{obs}}\mathcal{L}_{\mathrm{obs}}
+\lambda_{\mathrm{evt}}\mathcal{L}_{\mathrm{evt}}
+\lambda_{\mathrm{vqa}}\mathcal{L}_{\mathrm{vqa}},
\end{equation}
where, $\lambda_{\mathrm{obs}}$, $\lambda_{\mathrm{evt}}$, and $\lambda_{\mathrm{vqa}}$ are weighting coefficients that balance the contributions of the two objectives. Here, \(\mathcal{L}_{\mathrm{obs}}\) corresponds to \textit{unconscious learning} from naturally occurring visual transition.
\(\mathcal{L}_{\mathrm{evt}}\) and \(\mathcal{L}_{\mathrm{vqa}}\) correspond to \textit{conscious learning} through language-specified transitions and common sense. 
$\mathcal{L}_{\mathrm{evt}}$ uses the ground truth latent of a frame $v^l_{t+\Delta}$ in the adjacent event to perform teacher forcing on the predicted latent $\hat{v}^l_{t+\Delta}$ under the constraint of $e_{t+\Delta}$.
$\mathcal{L}_{\mathrm{vqa}}$ represents the standard VQA loss. 
Given a visual information $V$ and a question $l_q$, Orca learns to produce the target response $l_a$. The details of the sampling ratio, loss coefficients, and optimization settings are provided in \textbf{Appendix~\ref{sec:appendix_pretraining_settings}}.

\subsubsection{Pre-Training Data}
\label{sec:pretraining_data}

\begin{figure}[!htbp]
    \centering
    \includegraphics[width=1\linewidth]{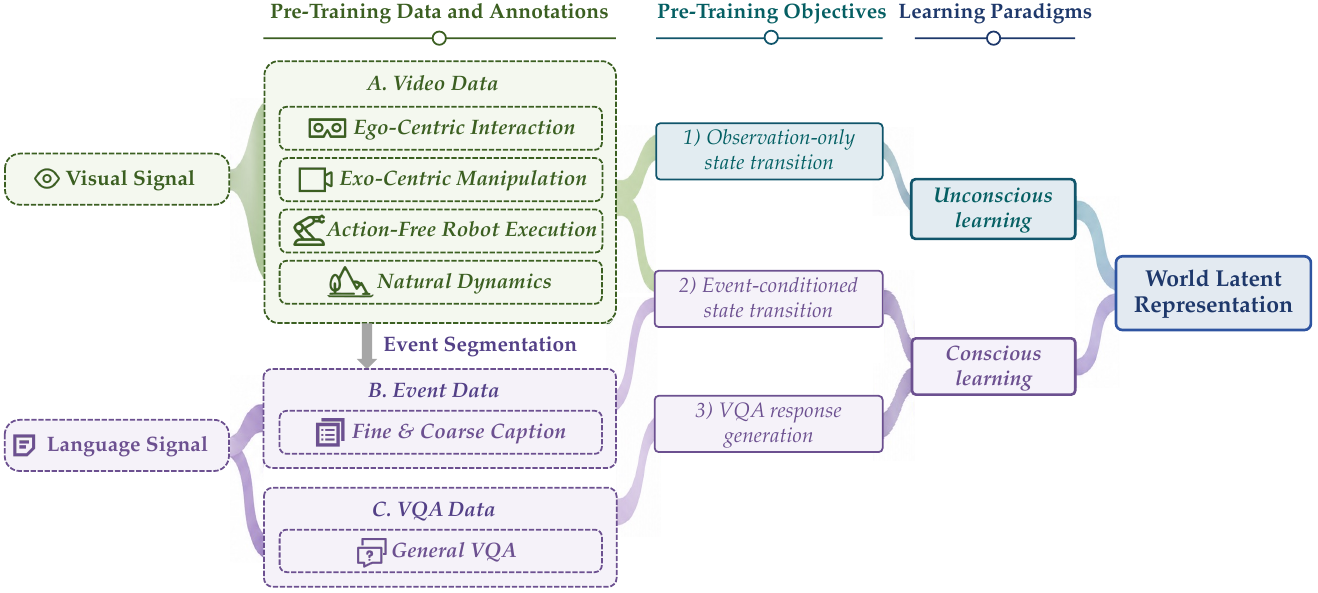} 
    \caption{\textbf{Overview of pre-training data.}
    Orca's pre-training data includes video, event, and VQA data. \textbf{\textit{A. Video Data}} supports \textit{1) Observation-only state transition}, \textbf{\textit{A. Video Data}} and \textbf{\textit{B. Event Data}} support \textit{2) Event-conditioned state transition}, and \textbf{\textit{C. VQA Data}} supports \textit{3) VQA response generation}.} %
    \label{fig:data_sankey}
\end{figure}

\paragraph{Data Organization.}
The three collections provide complementary supervision for learning world states and their transitions. The pre-training data is shown in Figure~\ref{fig:data_sankey}.

\vspace{-1ex}
\begin{enumerate}[
    leftmargin=1.35em,
    itemsep=0.35em,
    topsep=0.35em,
    parsep=0em
]
    \item[] 
    \textit{\textbf{A. Video Data}} is built from visual signals and covers four types of real-world observations: \textit{{ego-centric interaction}}, \textit{{exo-centric manipulation}}, {\textit{action-free robot execution}}, and {\textit{natural dynamics}}. {\textit{Ego-centric interaction}} captures first-views experience during physical interaction, {\textit{exo-centric manipulation}} provides third-views of object-centered changes, {\textit{action-free robot execution}} records embodied action in robotic environments, and {\textit{natural dynamics}} describes naturally evolving scenes. These data support \textit{1) observation-only state transition} and \textit{2) event-conditioned state transition}.

    \item[] 
    \textit{\textbf{B. Event data}} is derived from \textit{\textbf{A. Video Data}} through multi-level event segmentation and language annotation. {\textit{Coarse events}} describe the main steps of a temporal process, while {\textit{fine-grained events}} capture the shorter state transitions within each step. Each segmented event is paired with a caption that specifies the transition. This data supports \textit{2) event-conditioned state transition}.

    \item[]
    \textit{\textbf{C. VQA Data}} is constructed from language signals and video data, which teaches Orca to describe and interpret observed world states. This collection supports \textit{3) VQA response generation}.
\end{enumerate}
\vspace{-1ex}

Across these collections, data construction is grounded in the real world. The video data is built from real-world videos, while event and VQA data are constructed on top of these observations to describe state transition, physical relations, spatial configurations, behavioral intentions, and causal consequences. The existing data includes 125K hours of general video data, 160M of event annotations, and 11.5M of general VQA data. In this version, only one-tenth of the video data are used. The remaining data will be used in Orca's subsequent version iterations.

\subsection{Downstream Post-Training} \label{sec:downstream_readout}
\begin{figure}[!htbp]
    \centering
    \includegraphics[width=0.93\linewidth]{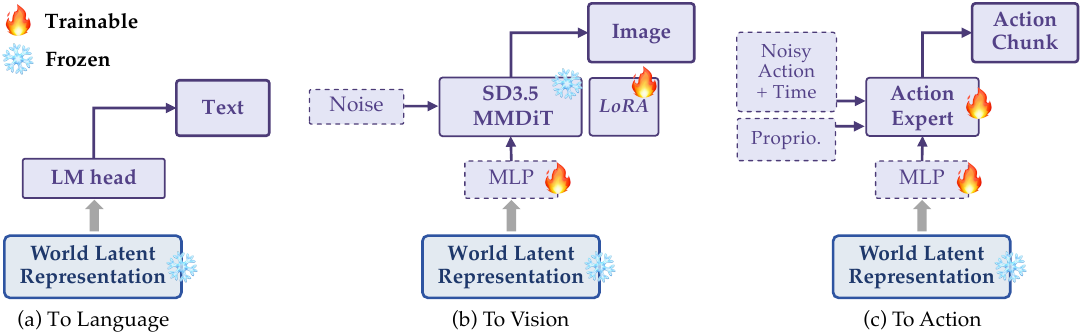}
    \caption{\textbf{Downstream readout architectures.} \textit{To language} reuses the \textit{LM head} for text readout. \textit{To vision} only trains an \textit{MLP adaptor} and \textit{LoRA} on top of a frozen \textit{SD3.5} to readout images. \textit{To action} trains an \textit{MLP adaptor} and a DiT-based \textit{Action Expert} from scratch. \textit{Action Expert} receives the latent, robot proprioception state, and noisy action to generate action chunks. The specific settings are shown in \textbf{Appendix \ref{sec:appendix_downstream_readout_settings}}.} \label{fig:downstream_readout_architecture}
\end{figure}

After pre-training, Orca is connected to downstream readout interfaces for language, vision, and action. \textbf{\textit{Note that our goal is to explore, in essence, whether the learned latent is effective for downstream tasks. So, Orca's backbone is always frozen, and only the corresponding readout modules are trainable.}} In other words, if Orca is intended to support vision tasks, it only trains the downstream image modules; if it is intended to serve embodied tasks, it only trains the action modules. Through these readouts, Orca exposes the latent as text, image, and action. Figure~\ref{fig:downstream_readout_architecture} provides an overview of the readout architectures. 

\subsubsection{To Language: Text Generation} \label{sec:downstream_language} 

As shown in Figure~\ref{fig:downstream_readout_architecture}(a).
Given a visual observation and an instruction, Orca produces the response through \textit{LM head}, without attaching an additional decoder. It expresses Orca's latent in natural language.

\subsubsection{To Vision: Image Prediction}
\label{sec:downstream_vision} 

\paragraph{Vision Readout Recipe.} 

As shown in Figure~\ref{fig:downstream_readout_architecture}(b), the vision readout maps the latent to a pixel-level image. Since Orca focuses on the encoder, the decoder uses a pre-trained model to show the effectiveness of latent.
The latent is passed through an \textit{MLP adaptor} and then used as a path input of a \textit{Stable Diffusion 3.5}~\citep{sd35}. The ground truth image with Gaussian noise is fed into another path of SD3.5 through a frozen VAE. The final predicted image is obtained through multi-step denoising. During this module training, only the \textit{MLP adaptor} and the \textit{LoRA} parameters are trainable. 

\paragraph{Image Prediction Data.}
This readout training uses paired current and target frames sampled from \textit{A. Video Data} in Figure \ref{fig:data_sankey}. Given the image and an instruction, Orca first produces the latent of the target frame using frozen Orca, and then the latent is input into the image readout module to obtain the image.

\subsubsection{To Action: Action Generation} 
\label{sec:downstream_action} 

\paragraph{Action Readout Recipe.} 
As shown in Figure~\ref{fig:downstream_readout_architecture}(c). 
The \textit{Action Expert} is a DiT-based model with flow-matching loss, and it is trained from scratch. \textit{Action Expert} receives the \textit{noisy action} with \textit{time embedding} and the \textit{proprioception}. The latent of $q_1$ is processed by the \textit{MLP adaptor} and input as a condition into the \textit{Action Expert}. Through multi-step denoising, the final \textit{Action Chunk} used to control the robot manipulation is obtained.
During this module training, only the \textit{MLP adaptor} and \textit{Action Expert} are trainable.

\paragraph{Embodied Action Data.} 
This readout training uses action-labeled data consisting of 5 tasks collected by the dual-arm wheeled humanoid robot. Note that \textit{Action Expert} has only seen 200 trajectories, instructions, visual information, and proprioception for each task. The details are shown in \textbf{Section \ref{sec:evaluation_toA}}.

\subsection{Infrastructures}
\label{sec:pretraining_infrastructures}

Orca's infra uses the self-developed FlagScale~\citep{flagscale} and makes the following improvements:

\vspace{-1ex}
\begin{enumerate}[
    leftmargin=1.35em,
    itemsep=0.35em,
    topsep=0.35em,
    parsep=0em
]
    \item[] \textbf{\textit{1) FlagScale training framework.}} 
    We use FlagScale and rebuild the Orca training with FSDP2, enabling more flexible parameter sharding, better memory control, and stable training.
    
    \item[] \textbf{\textit{2) Memory-efficient loss and recompute.}}
    We adopt Chunked Cross-Entropy Loss to avoid materializing full logits during loss computation, and further apply activation recomputation to trade moderate computation overhead for substantial memory savings, enabling larger batch sizes.
    
    \item[] \textbf{\textit{3) Communication scheduling.}}
    We introduce forward/backward pre-fetching to overlap FSDP all-gather communication with computation, and remove unnecessary FSDP sharding for visual blocks.
\end{enumerate}

With these optimizations, training throughput increases from 0.66 to 2.91 Samples/Sec/GPU, achieving approximately a 4.4$\times$ acceleration compared to the StarVLA \citep{StarVLA} commonly used in the embodied community. Optimization details and results are shown in \textbf{Appendix~\ref{sec:appendix_infrastructure}}.
\section{Evaluation}\label{sec:evaluation}
\setnavsection{sec:evaluation}

\subsection{Effectiveness and Scaling Behavior}
\label{sec:paradigm_effectiveness}

Before giving the downstream-specific results, we first explore whether Orca's core hypotheses hold. As presented in the introduction, Orca is designed to learn a world latent space through next state prediction, and this latent space is expected to support downstream readouts for understanding, prediction, and intervention. Therefore, we evaluate the Orca paradigm by answering two questions:

\begin{enumerate}[label={}, leftmargin=10pt, itemindent=0pt, itemsep=0.3em, topsep=0.3em]
    \item[] \textit{\textbf{$\cdot$ Question 1.1: Is Orca's learning paradigm effective as the model size and data scale up?}}
    \item[] \textit{\textbf{$\cdot$ Question 1.2: Can a stronger latent by pre-training improve downstream readout performance?}}
\end{enumerate}

\subsubsection{Loss of Proposed Learning Paradigm}
\label{sec:evaluation_scaling_loss}

\begin{wrapfigure}{r}{0.38\linewidth}
\vspace{-3.8em}
\centering
\includegraphics[width=\linewidth]{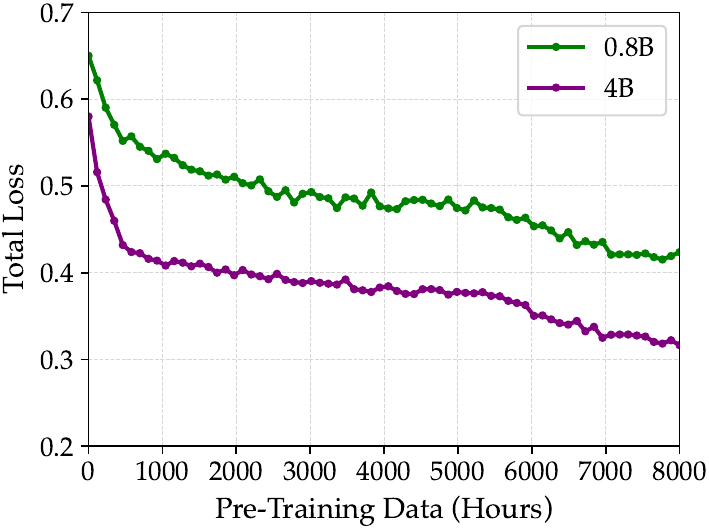}
\vspace{-2em}
\caption{\textbf{\small Loss of model and data scaling.}}\label{fig:scalinglaw_loss}
\vspace{-2em}
\end{wrapfigure}

To answer \textbf{\textit{Question 1.1}}, we first performed experiments with model sizes and data scaling, and the loss curves are shown in Figure \ref{fig:scalinglaw_loss}. The horizontal axis represents the amount of pre-trained video data (the unit is hours); the vertical axis represents the total loss, calculated by Equation \ref{eq:total_loss}. The green and the purple lines represent the trends of 0.8B and 4B model sizes, as the data scaling. The total loss has been on a downward trend.

Based on Figure~\ref{fig:scalinglaw_loss}, we obtained \textbf{\textit{Answer 1.1}}:

\begin{enumerate}[label={}, leftmargin=10pt, itemindent=0pt, itemsep=0.3em, topsep=0.3em]
\item[] \textit{\textbf{$\cdot$ Answer 1.1: Orca's learning paradigm is effective and scalable as the model size and data increase.}}
\end{enumerate}

The total loss of Orca decreases as the pre-training data scale up, and the larger Orca achieves a lower objective loss than the smaller one. This trend suggests that Orca provides an \textit{effective} learning paradigm for building world latent. 
Orca does not converge quickly, but rather continuously benefits from more data and larger model sizes. Its loss curve continues to show a significant downward trend, further demonstrating its \textit{scalability}.

\subsubsection{Relationship between latent and downstream readout performance}
\label{sec:evaluation_scaling_performance}

\begin{wrapfigure}{r}{0.7\linewidth}
\vspace{-1.3em}
\centering
\includegraphics[width=\linewidth]{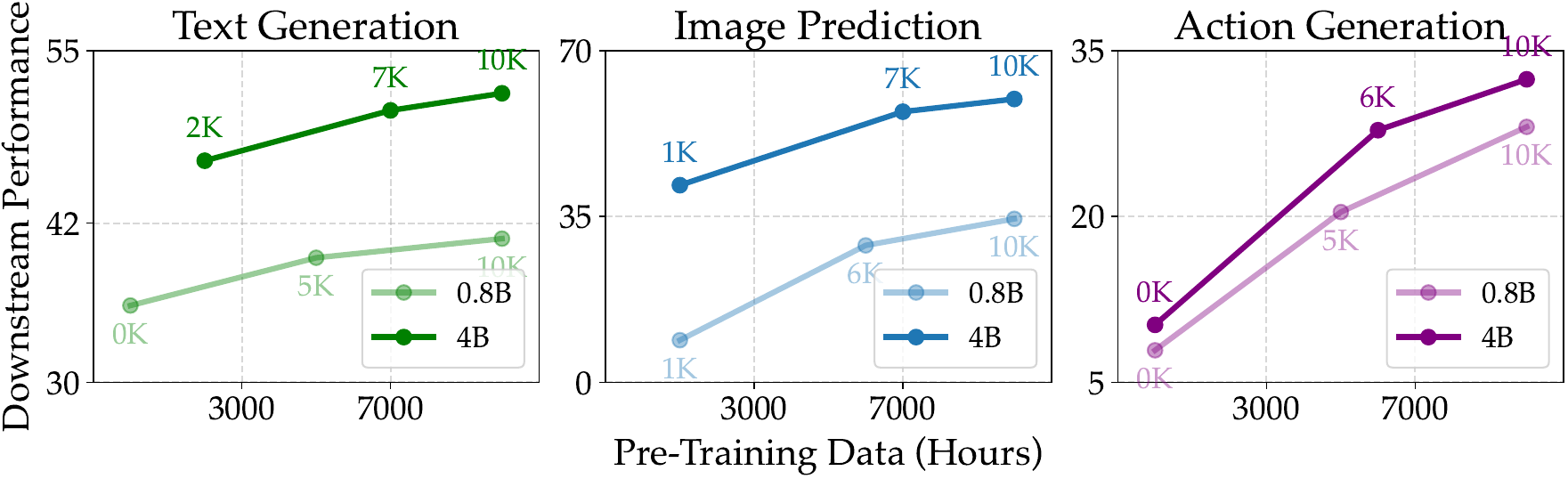}
\caption{\textbf{Scaling behavior on downstream readouts performance.}}\label{fig:scalinglaw_readout}
\vspace{-0.5em}
\end{wrapfigure}

To answer \textbf{\textit{Question 1.2}}, we performed probe experiments on Orca-0.8B and Orca-4B. We select some checkpoints from the pre-training process and apply them to downstream tasks to see if a strong world latent can lead to strong downstream readout performance. The readout performance curves are shown in Figure \ref{fig:scalinglaw_readout}.  Light-colored lines represent 0.8B, and dark-colored lines represent 4B. The corresponding points represent probes during pre-training. The horizontal axis is the amount of pre-trained video data, and the vertical axis is the corresponding readout performance.

The downstream readout performance across \textit{text generation}, \textit{image prediction}, and \textit{action generation}. The \textit{text generation} performance is the average performance on four benchmarks: TemporalBench, MVBench, SWITCH, and 3DRSBench. The details are shown in \textbf{Section \ref{sec:evaluation_toL}}. The \textit{image prediction} performance is the average performance of the proposed PRICE-V0.1 benchmark. The details are shown in \textbf{Section \ref{sec:evaluation_toV}}. Note that the above are all zero shots. The \textit{action generation} performance is the average performance on five real-robot out-of-domain tasks. The details are shown in \textbf{Section \ref{sec:evaluation_toA}}.

Based on Figure~\ref{fig:scalinglaw_readout}, we obtained \textbf{\textit{Answer 1.2}}:

\vspace{-1ex}
\begin{enumerate}[label={}, leftmargin=10pt, itemindent=0pt, itemsep=0.3em, topsep=0.3em]
\item[] \textit{\textbf{$\cdot$ Answer 1.2: Stronger world latent from pre-training leads to stronger downstream readouts.}} 
\end{enumerate}

Orca's backbone is frozen during readout post-training. As the pre-training data scale up, text, image, and action readouts all improve. This result shows that Orca's stronger latent space can enhance downstream readout performance. Surprisingly, no data with action labels was used in pre-training, but in \textit{action generation}, this paradigm brought gains by relying on video data. This emergent capability may alleviate, to some extent, the problem of low generalization caused by the scarcity of robot data.

\subsection{Downstream Readout Analysis}

Following the above discussion of \textbf{\textit{Question 1.2}} and \textbf{\textit{Answer 1.2}}, we present the quantitative evaluation results of Orca across three downstream tasks: \textit{text generation}, \textit{image prediction}, and \textit{action generation}. \textbf{\textit{Note that we do not construct or use any benchmark-specific training data for these evaluations, nor do we tune Orca on the evaluation benchmarks.}}

\vspace{-2ex}
\begin{enumerate}
    \item \textbf{\textit{Text Generation}}. It demonstrates the model's out-of-distribution (OOD) commonsense reasoning, comprehension capabilities, and high-level cognitive abilities.
    \item \textbf{\textit{Image Prediction}}. It visualizes this latent cognitive capability through OOD state transitions.
    \item \textit{\textbf{Action Generation}}. It executes the generated actions in real-world OOD scenarios, mapping the state transitions to action manipulation.
\end{enumerate}

\subsubsection{Comparison on Text Generation}\label{sec:evaluation_toL}

To evaluate how state transition modeling enhances abstract reasoning, we conducted text generation assessments on OOD understanding.
The details of benchmarks and baselines of the text generation can be seen in \textbf{Appendix \ref{sec:appendix_evaluation_toL}}.

\paragraph{Benchmarks.} 
We evaluate Orca on a complementary suite of benchmarks that probe different aspects of world-state and state-transition understanding: 
MVBench \citep{MVBench}, TemporalBench \citep{cai2024temporalbench}, 3DSRBench \citep{ma20253dsrbench}, and SWITCH \citep{switch2025}.

\paragraph{Baselines.} We compare Orca with two categories of baselines:

\vspace{-2ex}
\begin{itemize}
    \item \textit{\textbf{World models}}: V-JEPA 2.1~\citep{V-JEPA-2.1}, Emu3~\citep{Emu3}, and Emu3.5~\citep{Emu35}.
    \item \textit{\textbf{Vision-language models}}: Qwen3.5~\citep{qwen35}, Gemma 4~\citep{Gemma4}, DeepSeek-VL2~\citep{DeepSeek-VL2}, MiniCPM-V-4.6~\citep{MiniCPM-o-4.5}, and SmolVLM2~\citep{SmolVLM}.
\end{itemize}

\paragraph{Results and Analysis.} 
Based on Table~\ref{tab:comparison_tol}, Orca achieves the best overall result among the same-size VLMs and the large-size world models, demonstrating the advantages of the proposed learning paradigm.

\begin{table}[!htbp]
  \centering
  \small
  \begin{threeparttable}
  \caption{\textbf{The comparison of the text generation.} $\uparrow$ represents the higher value, the better.}
  \label{tab:comparison_tol}
    \begin{tabular}{lc|cccc|c}
    \toprule[1pt]
    Model & Size (B) & MVBench $\uparrow$ & TemporalBench $\uparrow$ & 3DSRBench $\uparrow$ & SWITCH $\uparrow$ & \textbf{Avg.} \\
    \midrule[0.5pt]
    \rowcolor[rgb]{ .949,  .949,  .949}\multicolumn{7}{l}{\textit{World Models (Large size)}} \\

    V-JEPA 2.1\tnote{1} \ (+LLaMA3-8B) & 10 & 75.4 & 28.5 & / & / & / \\
    Emu3\tnote{2} & 8 & 35.2 & 9.5 & 39.1 & 38.0 & 30.4 \\
    Emu3.5 & 34 & 39.5 & 9.5 & 31.3 & 38.9 & 29.8 \\

    \midrule[0.5pt]
    \rowcolor[rgb]{ .949,  .949,  .949} \multicolumn{7}{l}{\textit{Vision Language Models (Tiny size)}} \\
    Qwen3.5 & 0.8 & 52.7 & 19.1 & 21.8 & 38.8 & 33.1 \\
    Gemma 4 & 2 & 32.5 & 17.1 & 29.5 & 39.9 & 29.8 \\
    SmolVLM2 & 2 & 48.7 & 18.4 & 35.5 & 32.0 & 33.7 \\
    MiniCPM-V-4.6 & 2 & 41.4 & 21.2 & 47.7 & 41.2 & 37.9 \\

    \midrule[0.5pt]
    \rowcolor[rgb]{ .949,  .949,  .949} \multicolumn{7}{l}{\textit{Vision Language Models (Small size)}} \\
    DeepSeek-VL2 & 3 & 40.5 & 21.0 & 32.1 & 35.5 & 32.3 \\
    Gemma 4 & 4 & 45.6 & 20.2 & 44.8 & 52.4 & 40.8 \\
    Qwen3.5 & 4 & 67.1 & 25.2 & 48.1 & 46.5 & 46.7 \\

    \midrule[0.5pt]
    \rowcolor[rgb]{ .89,  .949,  .851}
    & 0.8 & 53.6 & 22.6 & 43.4 & 43.7 & 40.8 \\
    \rowcolor[rgb]{ .89,  .949,  .851}
    \multirow{-2}{*}{\textbf{Orca}} & 4 & 65.3 & 34.2 & 52.1 & 55.6 & \textbf{51.8} \\
    \bottomrule[1pt]
    \end{tabular}

    \begin{tablenotes}
    \footnotesize
    \item[1] Since V-JEPA 2.1 does not publicly disclose the alignment data and adjusted LLaMA3-8B weights, its results are taken from the original paper \citep{V-JEPA-2.1}.
    \item[2] Emu3 denotes the \textit{Emu3-Chat} version.
    \end{tablenotes}
  \end{threeparttable}
\end{table}

We believe that a key characteristic of world models lies in their ability to construct a unified latent space for the physical world. Such models should not only be able to internalize the natural evolution of environmental dynamics over time, but also accurately predict complex state transitions caused by external interventions and behaviors. 
We also evaluated the capability dimensions to explore the boundaries. Specifically, we identified four core capability dimensions:

\vspace{-1ex}
\begin{enumerate}[
    leftmargin=1.35em,
    itemsep=0.35em,
    topsep=0.35em,
    parsep=0em
]
    \item[] \textbf{\textit{1) State Transition.}} 
    It focuses on state transitions induced by actions or temporal evolution. 
    
    \item[] \textbf{\textit{2) Commonsense Reasoning.}}
    It evaluates the internalization of social and physical commonsense knowledge, as well as the ability to reason causally. 
    
    \item[] \textbf{\textit{3) Spatial Relations.}}
    It measures the understanding of three-dimensional geometric relationships. 

    \item[] \textbf{\textit{4) Dynamic Motion.}}
    It assesses quantitative reasoning over kinematic properties, including velocity, direction vectors, and higher-order motion characteristics. 
\end{enumerate}

\begin{table}[!htbp]
  \centering
  \small
  \begin{threeparttable}
  \caption{The cross-benchmark general capability comparison of the text generation.}
  \label{tab:comparison_tol_cap}
    \begin{tabular}{l|cccc} 
    \toprule[1pt]
    Model  & State Transition\tnote{1} & Commonsense Reasoning\tnote{2} & Spatial Relations\tnote{3} & Dynamic Motion\tnote{4} \\
    \midrule[0.5pt]

    Qwen3.5-4B & 51.86 & 57.76 & 54.68 & 57.03 \\

    \rowcolor[rgb]{ .89,  .949,  .851}
    \textbf{Orca-4B}
    & \textbf{64.13} {\scriptsize(+12.27\%)}  
    & \textbf{62.95} {\scriptsize(+5.19\%)}  
    & \textbf{55.25} {\scriptsize(+0.57\%)}  
    & \textbf{65.55} {\scriptsize(+8.52\%)}  \\
    \bottomrule[1pt]
    \end{tabular}

    \begin{tablenotes}
    \footnotesize
    \item[1] 644 samples from {MVBench and SWITCH}.
    \item[2] 1,676 samples from {MVBench and SWITCH}.
    \item[3] 11,686 samples from 3DSRBench.
    \item[4] 1,736 samples from {TemporalBench and MVBench}.
    \end{tablenotes}
  \end{threeparttable}
\end{table}

By aggregating samples associated with each capability dimension across multiple benchmarks and computing the corresponding average success rates, we obtain a large-scale and benchmark-agnostic evaluation framework, as summarized in Table~\ref{tab:comparison_tol_cap}. The results of Table \ref{tab:comparison_tol_cap} demonstrate that:

\vspace{-1ex}
\begin{enumerate}[
    leftmargin=1.35em,
    itemsep=0.35em,
    topsep=0.35em,
    parsep=0em
]
    \item[] \textbf{\textit{1) More accurate state transition.}} 
    Orca predicts future states more accurately and demonstrates a deeper understanding of temporal dynamics.
    
    \item[] \textbf{\textit{2) Common-sense and counterfactual reasoning.}}
    Orca achieves more reliable common-sense reasoning and counterfactual reasoning through causal alignment of conscious learning.
    
    \item[] \textbf{\textit{3) Strong spatial understanding.}}
    Orca can capture geometric continuity, reduce spatial inconsistencies, and improve robustness under complex perspectives.

    \item[] \textbf{\textit{4) Dynamic motion consistency.}}
    Orca can better capture temporal continuity and motion inertia.
\end{enumerate}

\subsubsection{Comparison on Image Prediction}\label{sec:evaluation_toV}

To visualize the capability of the state transition, we performed a comparison on image prediction. The details of benchmarks and baselines of the image prediction can be seen in \textbf{Appendix \ref{sec:appendix_evaluation_toV}}.

\paragraph{Benchmark.} Our motivation is not to create a \textit{painter}, but to explore whether the latent possesses the ability to predict future states. So, instead of generating or simulating scenarios, we build a real-world dataset, PRICE-V0.1 (i.e., Prediction of Real-world Interactions with Constraints Evaluation). 
PRICE-V0.1  benchmark is shown in \textbf{Appendix \ref{sec:appendix_evaluation_toV_benchmark}}. 

\paragraph{Metrics.} In PRICE-V0.1, we use Gemini 3.1 Pro \citep{Gemini-3.1-pro}, GPT 5.4 \citep{GPT54}, Doubao-Seed-2.0-Pro-260215 \citep{doubao-seed-2.0-pro}, and open-source Gemma 4-31B \citep{Gemma4} for evaluation. The specific evaluation prompt is shown in the \textbf{Listing~\ref{lst:evaluator_prompt}}. 

\paragraph{Baselines.} We selected recent image generation models with \textit{\textbf{a similar size}} to Orca as baselines: including OmniGen2 \citep{OmniGen2} (3B VLM + 4B vision decoder), FLUX.1-Kontext \citep{FLUX.1-Kontext} (12B vision decoder), and Flux.2 [klein] \citep{flux2} (4B VLM + 4B vision decoder). 

\begin{figure}[!htbp]
    \centering
    \includegraphics[width=1\linewidth]{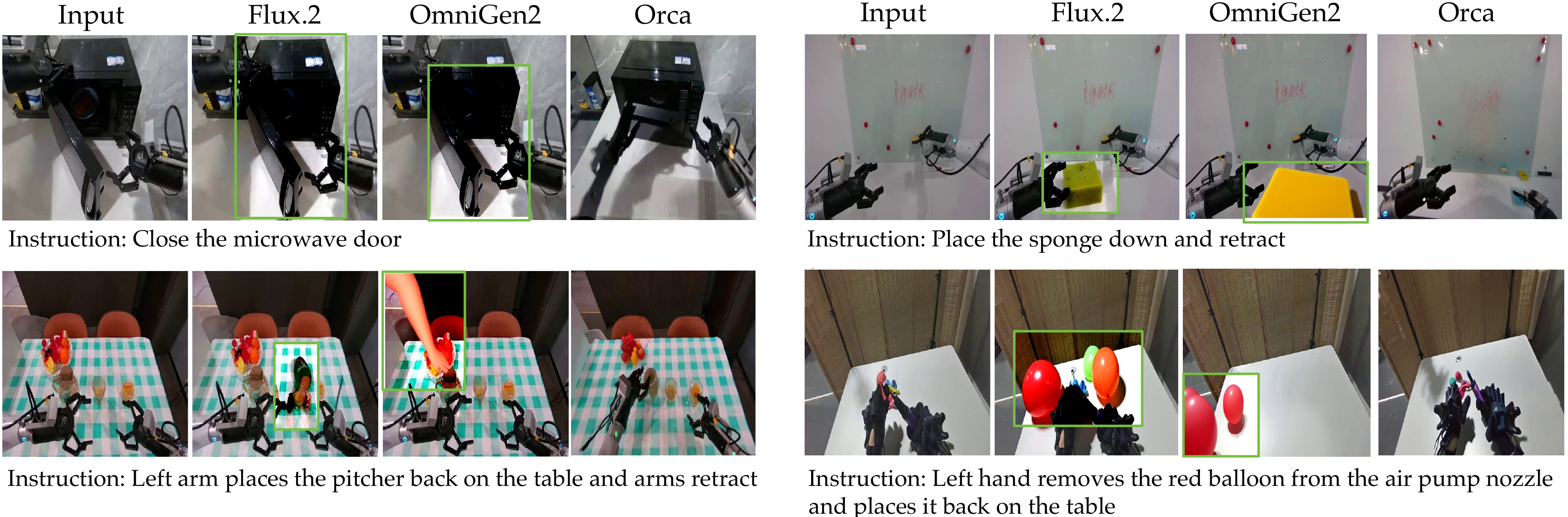} 
    \caption{\textbf{Visual comparison of image prediction in the real world.}}
    \label{fig:tov_qualitative_main}
    \vspace{-0.3em}
\end{figure}

\paragraph{Results and Analysis.} 
Based on Table~\ref{tab:toV_results} and Figure~\ref{fig:tov_qualitative_main}, we obtained two conclusions:

\begin{table}[!htbp]
  \centering
  \small
  \caption{The comparison of the PRICE-V0.1. $\uparrow$ represents the higher value, the better. In \textbf{Avg.}, \textit{a}$\pm$\textit{b} is \textit{avg}$\pm$\textit{std}. A larger \textit{avg} and a smaller \textit{std} value represent a better result. \textbf{Bold} represents the best value.}\label{tab:toV_results}%
    \begin{tabular}{cc|cccc|c}
    \toprule[1pt]
    Model&Size (B)& Gemini 3.1 Pro $\uparrow$ & GPT 5.4 $\uparrow$& Doubao-Seed-2.0 $\uparrow$& Gemma 4-31B $\uparrow$& \textbf{Avg.} \\
    \midrule[0.5pt]
    OmniGen2 &3+4& 24.6	&46.8	&41.4	&45.5	&39.6{\scriptsize$\pm$10.2} \\
    FLUX.1-Kontext  &12& 21.6&	46.9	&42.7	&52.5	&40.9{\scriptsize$\pm$13.5} \\
    FLUX.2 [klein] &4+4& 29.7&	64.6	&60.0&	70.2	&56.1{\scriptsize$\pm$18.1} \\
    \midrule[0.5pt]
    \rowcolor[rgb]{ .89,  .949,  .851} &0.8+2& 17.0 &	48.5 	& 46.0 	& 26.5&34.5{\scriptsize$\pm$15.3} \\
    \rowcolor[rgb]{ .89,  .949,  .851}\multirow{-2}{*}{\textbf{Orca}} &4+2& 44.0& 	67.9	& 61.0	& 66.3& 	\textbf{59.8{\scriptsize$\pm$10.9}} \\
    \bottomrule[1pt]
    \end{tabular}%
\end{table}

\begin{enumerate}[
    leftmargin=1.35em,
    itemsep=0.35em,
    topsep=0.35em,
    parsep=0em
]
    \item[] \textit{\textbf{1) Orca's learned world latent transfers effectively to image readout.}} 
    Compared with recent image generation baselines, Orca achieves the best average performance on PRICE and remains competitive across different real-world interaction sources. 
    This indicates that the learned world latent contains predictive information about future visual states under real-world interactions.

    \item[] \textit{\textbf{2) Orca better predicts interaction-conditioned state changes.}} 
    As shown in Figure~\ref{fig:tov_qualitative_main}, general image generation baselines suffer from typical flaws, such as the appearance or teleportation of irrelevant objects, hallucinatory human hands, poor instruction adherence, and biases stemming from prior knowledge. Orca better preserves robot morphology, scene and object consistency, contact relationships, and instruction following.
\end{enumerate}

These results suggest that the learned world latent provides useful state-transition information for visual readout, enabling more physically grounded image prediction for real-world interactions.

\subsubsection{Comparison on Action Generation}\label{sec:evaluation_toA}
To truly apply state transition modeling capabilities to the real world, we performed the embodied real-robot tasks. The details of benchmarks, metrics, and baselines are shown in \textbf{Appendix \ref{sec:appendix_evaluation_toA}}.

\paragraph{Benchmarks.} 

We used a dual-arm wheeled robot to collect data on five tasks: \textit{Take Book}, \textit{Stacked Bowls}, \textit{Pull Out Tissue}, \textit{Stamp}, and \textit{Scoop Sugar}. We performed two OOD settings: \textit{environment} and \textit{object OOD}. 

\paragraph{Metrics.} 
We report the rule-based scores, which measure key-stage task completion. The rule-based score is shown in Table \ref{tab:real_robot_scoring}.
We further employ PRM-as-a-Judge series~\citep{PRM-as-a-Judge, judge1.5} to provide dense trajectory-level diagnostics.

\paragraph{Baselines.} 
We compare Orca with V-JEPA 2.1~\citep{V-JEPA-2.1}, Qwen3.5~\citep{qwen35}, and $\pi_{0.5}$~\citep{pi0.5}. 
For V-JEPA 2.1 and Qwen3.5, we connect them to the same \textit{Action Expert} as Orca, i.e., V-JEPA 2.1 w/ AE and Qwen3.5 w/ AE, so that the comparison reflects the quality of the learned latent for action readout. V-JEPA 2.1 uses the latent as conditions, and Qwen3.5 uses the last hidden state as conditions. The remaining inputs of \textit{Action Expert} and training steps are consistent with Orca.
We also compare with $\pi_{0.5}$ as a strong VLA baseline pre-trained on large-scale robot data.

\begin{table}[t]
  \centering
  \small
  \setlength{\tabcolsep}{2.7mm}
  \begin{threeparttable}
  \caption{\textbf{Comparison on action generation.} $\uparrow$ represents the higher value, the better, and vice versa.
  The metrics are trajectory-level diagnostics from PRM-as-a-Judge. \textbf{Note} that the backbones of all methods are frozen, and \textit{Action Experts} for Orca, V-JEPA 2.1, and Qwen3.5 are trainable from scratch.}\label{tab:comparison_changing_world}
    \begin{tabular}{c|c|cccccccc}
    \toprule[1pt]
    \textbf{\textit{Environment OOD}}& Rule-based $\uparrow$ & M25 $\uparrow$\tnote{1} & M50 $\uparrow$\tnote{1} & SR $\uparrow$\tnote{2} & MaxP-F $\uparrow$\tnote{3} &  FNS $\uparrow$\tnote{4} & DRR $\uparrow$\tnote{5}  & SQS $\uparrow$\tnote{6} \\
    \midrule[0.5pt]
    V-JEPA 2.1 &15.2  & 40    & 12    & 0     & 23.0   & 13.9   & 25.8  &  0.0\\
    Qwen3.5 & 12.4  & 26    & 10    & 0     & 18.3   & 11.2   & 19.2  &  0.0\\
    $\pi_{0.5}$ & 27.6  & 54    & \textbf{16} & 2     & 27.9  & 17.7   & 31.5  &  1.5\\
    \rowcolor[rgb]{ .89,  .949,  .851}\textbf{Orca} &\textbf{36.6} & \textbf{64} & \textbf{16} & \textbf{4} & \textbf{33.9}   & \textbf{19.3}  & \textbf{32.9} & \textbf{1.8} \\
    \midrule[1pt]
    
    \multicolumn{1}{c|}{\textbf{\textit{Object OOD}}} & Rule-based $\uparrow$   & M25 $\uparrow$ & M50 $\uparrow$ & SR $\uparrow$ & MaxP-F $\uparrow$ & FNS $\uparrow$  & DRR $\uparrow$  & SQS $\uparrow$ \\
    \midrule[0.5pt]
    V-JEPA 2.1 &18.8  & 14    & 2     & 0     & 11.8  & 6.3    & 15.2  & 0.0 \\
    Qwen3.5 & 8.6   & 10    & 0     & 0     & 7.9   & 4.0     & 4.61  & 0.0\\
    $\pi_{0.5}$& \textbf{31.2} & \textbf{54}    & \textbf{12} & \textbf{8} & \textbf{25.1}  & \textbf{12.9}   & 21.9  & \textbf{4.5} \\
    \rowcolor[rgb]{ .89,  .949,  .851}\textbf{Orca} & 28.2  & 46 & \textbf{12}    & \textbf{8} & 21.8  &  10.8 & \textbf{27.7} & 3.9 \\
    \midrule[1pt]
    
    \multicolumn{1}{c|}{\textbf{\textit{Overall}}} &  Rule-based $\uparrow$   & M25 $\uparrow$ & M50 $\uparrow$ & SR $\uparrow$ & MaxP-F $\uparrow$ & FNS $\uparrow$ & DRR $\uparrow$  & SQS $\uparrow$ \\
    \midrule[0.5pt]
    V-JEPA 2.1 & 17.0  & 27    & 7     & 0     & 17.4  & 10.1    & 20.5  & 0.0 \\
    Qwen3.5 & 10.5  & 18    & 5     & 0     & 13.1  & 7.6     & 11.9  & 0.0 \\
    $\pi_{0.5}$  & 29.4  & 54    & \textbf{14} & 5     & 26.5  & \textbf{15.3}   & 26.7  & \textbf{3.0} \\
    \rowcolor[rgb]{ .89,  .949,  .851}\textbf{Orca} & \textbf{32.4} & \textbf{55} & \textbf{14}    & \textbf{6} & \textbf{27.9} & 15.1 & \textbf{30.3} & 2.9 \\
    \bottomrule[1pt]
    \end{tabular}%
    \begin{tablenotes}
        \footnotesize
        \item[1] M25 and M50 are Milestone25\% and Milestone50\%. They are the proportions of the trajectory reaching 25\% and 50\%.
        \item[2] SR is the binary Success Rate. The unit is \%.
        \item[3] MaxP-F is MaxProcess in Failure. It represents the max-level execution process in the failure.
        \item[4] FNS is Failure Near-Success Score. It measures the progress achieved by failed trajectories before termination.
        \item[5] DRR is the Drawdown Recovery Ratio. It measures recovery after the largest progress drawdown.
        \item[6] SQS is the Success Quality Score. It measures the stability, smoothness, and high quality in the success process.
    \end{tablenotes}
  \end{threeparttable}
\end{table}%

\begin{figure}[!htbp]
    \centering
    \includegraphics[width=1\linewidth]{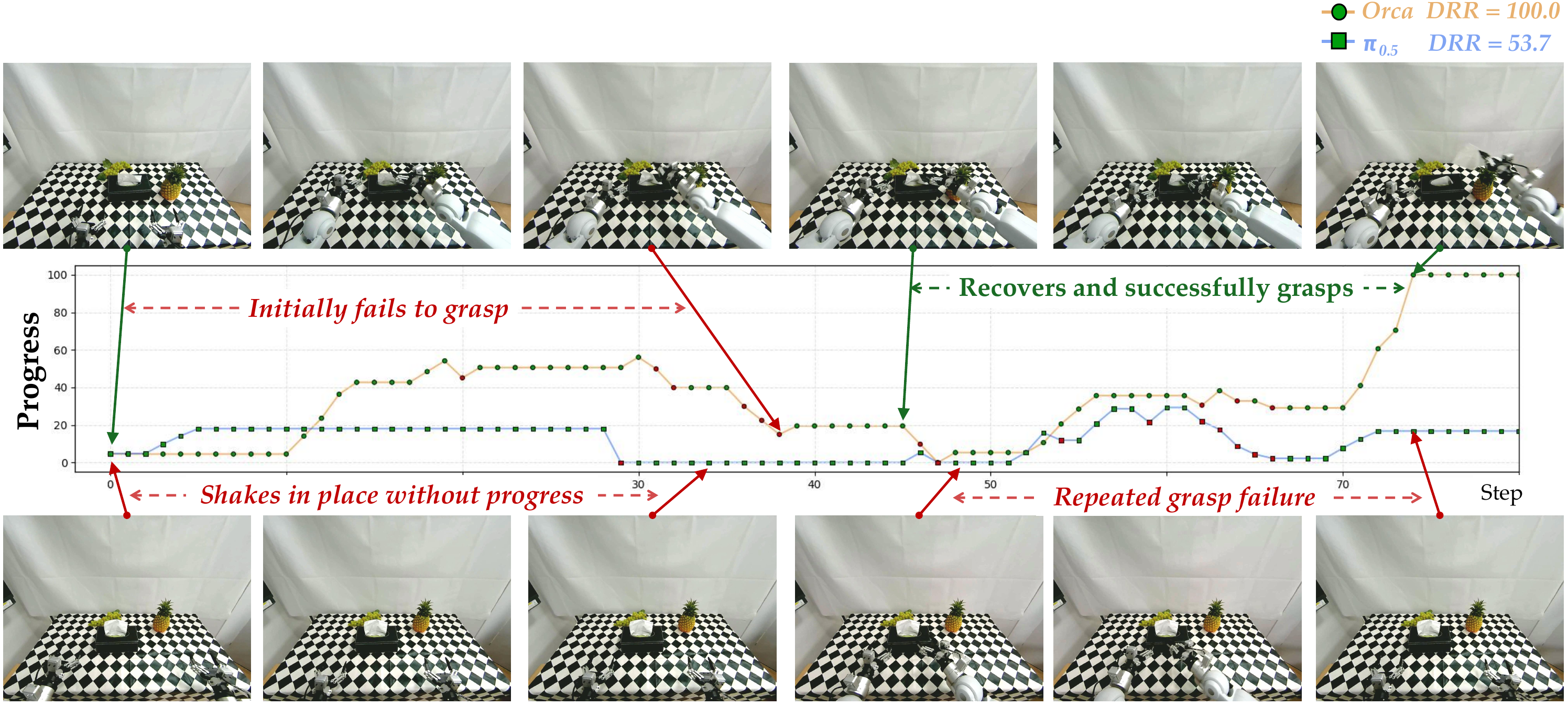}
\caption{
\textbf{Recovery after repeated grasp failures.}
Orca recovers from early spoon-grasp failures and eventually makes progress, while $\pi_{0.5}$ remains unstable with repeated failed attempts.
}
\label{fig:recovery_after_failure}
\end{figure}

\paragraph{Results and Analysis.} 
Based on Table~\ref{tab:comparison_changing_world}, we obtained two conclusions:

\begin{enumerate}[
    leftmargin=1.35em,
    itemsep=0.35em,
    topsep=0.35em,
    parsep=0em
]
    \item[] \textit{\textbf{1) Orca's learning paradigm and learned world latent transfers effectively to action readout.}} 
    Under the from-scratch Action Expert, 
    Orca outperforms Qwen3.5 in all OOD settings, achieving a breakthrough from 0\% success rate. It is also comparable to the powerful pre-trained $\pi_{0.5}$. This indicates that Orca's learning paradigm has a significant effect on action generation.

    \item[] \textit{\textbf{2) \textbf{Orca consistently advances the task and recovers better from execution errors.}}} The metrics show that Orca is more likely to make meaningful intermediate progress during execution, while suffering less from stagnation. Its higher FNS indicates that even when a trajectory eventually fails, Orca can reach later task stages before termination. Its higher DRR suggests that Orca is better at correcting deviations and continuing the task after progress drops. Figure~\ref{fig:recovery_after_failure} provides a qualitative case.
\end{enumerate}

These results suggest that Orca produces action trajectories that move further, get stuck less often, and recover more effectively after mistakes.

\vspace{-1em}
\subsection{Ablation}\label{sec:exploration}

In the current learning paradigm, there are three losses, i.e., \textit{1) Observation-only state transition} $\lambda_{\mathrm{obs}}$; \textit{2) Event-conditioned state transition} $\lambda_{\mathrm{evt}}$; \textit{3) VQA response generation} $\lambda_{\mathrm{vqa}}$. So we ablated the different losses. The ablation results are shown in Table \ref{tab:ablation}. The results in Table \ref{tab:ablation} demonstrate that:

\begin{table}[!htbp]
  \centering 
\caption{\textbf{Ablation results.} ``-'' means the setting does not work. The first three rows average two metrics, while the last two average all three. Text, image, and action denote the average scores on text benchmarks, PRICE-V0.1, and overall rule-based action evaluation.}\label{tab:ablation}%
    \begin{tabular}{ccc|ccc|c}
    \toprule[1pt]
     $\lambda_{\mathrm{obs}}$   & $\lambda_{\mathrm{evt}}$ & $\lambda_{\mathrm{vqa}}$ & Text Generation & 
     Image Prediction& Action Generation & \textbf{Average} \\
     \midrule[0.5pt]
     &    & \ding{51}  & 48.4 & - & 10.2 & 29.3 \\
     
     \ding{51}    & \ding{51}  &  &    -  & 58.2  & 30.9  & 44.6 \\
    \ding{51}&     &   \ding{51}  & 50.5 & - & \textbf{32.6} & 41.6 \\
    \midrule[0.5pt]
    &  \ding{51}  & \ding{51}    & 50.1 & 54.7 & 23.0 & 42.6 \\
    \rowcolor[rgb]{ .89,  .949,  .851}\ding{51}& \ding{51}    & \ding{51}    & \textbf{51.8} & \textbf{59.8} & 32.4& \textbf{48.0} \\
    \bottomrule[1pt]
    \end{tabular}%
    \vspace{-1em}
\end{table}%

\begin{enumerate}[
    leftmargin=1.35em,
    itemsep=0.35em,
    topsep=0.35em,
    parsep=0em
]
\item[] \textit{\textbf{1) The three pre-training objectives provide the most balanced downstream readouts.}}
When $\lambda_{\mathrm{obs}}$, $\lambda_{\mathrm{evt}}$, and $\lambda_{\mathrm{vqa}}$ are jointly used, Orca achieves the most balanced performance across text, image, and action readouts. This shows that the three objectives jointly constrain the world latent space from natural dynamics, semantic conditions, and language supervision.

\item[] \textit{\textbf{2) Observation-only transition is especially important for action readout.}} 
Adding $\lambda_{\mathrm{obs}}$ clearly improves action generation. This suggests that dense natural dynamics from continuous videos provide useful information about temporal changes, object motion, and local physical interactions, which are critical for real-robot action generation.

\item[] \textit{\textbf{3) Event-conditioned transition is the key supervision for vision readout.}} 
Image prediction requires the model to infer a target state under a semantic condition. $\lambda_{\mathrm{evt}}$ aligns language-described events with visual state transition, enabling Orca to predict instruction- or event-guided target states rather than only modeling unconditional visual dynamics.

\item[] \textit{\textbf{4) VQA response generation preserves the language interface and strengthens semantic grounding.}} 
$\lambda_{\mathrm{vqa}}$ enables Orca to maintain natural-language readout ability and provides semantic and commonsense constraints for the learned world latent space. When combined with the two state-transition objectives, it further improves the overall balance among different downstream readouts.
\end{enumerate}

\vspace{-1.5em}
\section{Conclusion}
\label{sec:conclusion}
\setnavsection{sec:conclusion}
We presented Orca, a world learner built around a world latent space. 
Rather than being purpose-built for isolated downstream tasks such as question answering, visual frame prediction, or action generation, Orca adopts a fundamentally different modeling paradigm:
It first learns an internal representation of world states from multimodal world signals, and subsequently exposes this representation via a suite of dedicated readout interfaces.
This design shifts the modeling target from next-token/frame/action prediction toward next-state prediction. 
Taken together, Orca constitutes an early exploratory milestone on the path toward building general-purpose world foundation models.

\vspace{-0.5em}
\paragraph{Discussion \& Limitation.}
Orca is still an early step toward general world foundation models. We discuss its current boundaries together with the research directions they suggest for the community.

\vspace{-1ex}
\begin{enumerate}[
    leftmargin=1.35em,
    itemsep=0.35em,
    topsep=0.35em,
    parsep=0em
]

    \item[] \textit{\textbf{1) Limited multimodal world signals.}}
    Orca currently learns mainly from vision and language, which cover only a subset of the multimodal world signals. However, many state transitions are expressed through other sensory or physical signals. For example, whether water is boiling can often be inferred from sound before clear visual changes appear, and tactile or force feedback can reveal contact, slippage, stiffness, or whether an object is firmly grasped. Future world learners should incorporate richer neural and physical signals, such as audio, tactile, force, light, and proprioception, and eventually extend to broader scientific domains to build a more complete world latent space.

    \item[] \textit{\textbf{2) ViT space supervision.}} Orca aimed to provide a new learning paradigm that, in all other respects, employed a naive setup, thus using a pre-trained VLM and supervising visual state prediction within a frozen vision encoder. This design simplified the training process. However, this also aligns the learned state space with the semantic space. A general world foundation model should learn a unified world space directly from multi-source signals. These signals should jointly define and constrain the state, rather than relying on any single pre-trained modality space as the supervision target.

    \item[] \textit{\textbf{3) Model size limited.}} Due to resource constraints, our current experiments are mainly conducted at the 4B and 0.8B scale. The current scale is insufficient to fully integrate greater world knowledge, more modalities, and more data. We found that the 4B model exhibits a trade-off among language, image, and action readout performance as pre-training progresses, and this trade-off is more pronounced in the 0.8B model. Therefore, although we have created 125K hours of video data and 160M event annotations, the current training only uses one-tenth of the inventory data. This indicates that world learning is not only limited by the data scale, but also requires sufficient model capacity.

    \item[] \textit{\textbf{4) Vision benchmark limited.}} 
    Although the proposed PRICE-V0.1 covers multiple real-world data sources, its scale, diversity, and interaction richness are still limited. We hope it can serve as an initial step toward a more comprehensive evaluation of real-world state prediction.

    \item[] \textit{\textbf{5) Short-horizon transition supervision.}}
    The current state-transition supervision is constrained by the event annotation. Most event annotations describe short-horizon, minute-level state transitions, which are suitable for learning local transitions but insufficient for modeling long-term state evolution over hours, days, or even longer horizons.

    \item[] \textit{\textbf{6) Downstream readout limited.}} Currently, we have verified that the world latent we have learned is readout language, vision, and action. However, this is far from enough, as information from other fields such as hearing, quantum circuits, and proteins remains an important part of the world. 
    
    \item[] \textit{\textbf{7) Loss function limited.}} We use three losses to fully train Orca, but this is not consistent enough for the Next-State-Prediction modeling. A simpler loss and supervision need to be proposed.

    \item[] \textit{\textbf{8) Embodied task difficulty limited.}} Our settings are quite stringent, resulting in lower performance. However, it's undeniable that the current embodiment tasks are still relatively short and easy.

\end{enumerate}

\vspace{-0.5em}
\paragraph{Future Works.}
We also provided some inspiration for the community, including:

\vspace{-1ex}
\begin{enumerate}[
    leftmargin=1.35em,
    itemsep=0.35em,
    topsep=0.35em,
    parsep=0em
]
    \item[] \textit{\textbf{1) More modalities input.}} The crucial next step is not simply to add more modalities, but to align them to the same underlying state to better constrain state transitions with the laws of physics.

    \item[] \textit{\textbf{2) Toward native world-state modeling.}} Native world foundation models can be pre-trained from scratch. To overcome the constraints imposed by a certain existing ViT space or other embedding model spaces, a unified world latent space can be learned directly from multi-source world signals, and a native world model can be trained from scratch.
    
    \item[] \textit{\textbf{3) A world model state transition evaluation system.}} This system constructs a unified evaluation framework for state prediction, intervention response, physical quantifiability, and counterfactual inference, preventing world models from remaining solely at the visual generation level. 

    \item[] \textit{\textbf{4) Model-Data-Evaluation self-evolutionary closed loop.}} The model autonomously generates interaction trajectories and counterfactual samples, which are automatically evaluated and value-filtered before being fed back into the training system, forming a self-evolutionary closed loop of ``data generation—data filtering-training—leap''.

    \item[] \textit{\textbf{5) Expanding the boundaries of human cognition.}} Gradually extending from embodied intelligence to complex systems such as AI for science, microscopic quantum mechanics, macroscopic universe, and life sciences, using a unified state transition world representation to support scientific discovery and the expansion of cognitive boundaries.

\end{enumerate}
\section{Author List}\label{sec:contributors}
\setnavsection{sec:contributors}

\subsection{Core Contributors} 
Yihao~Wang\textsuperscript{*}, \
Yuheng~Ji\textsuperscript{*,$\dagger$}, \ 
Mingyu~Cao\textsuperscript{*}, \
Yanqing~Shen\textsuperscript{*}, \
Runze~Xiao\textsuperscript{*}

\paragraph{Model Pre-Training.} 
Huaihai~Lyu, \
Senwei~Xie, \
Mingyu~Cao\textsuperscript{*}

\paragraph{Data Infra.}
Euan~Liu, \
Klara~Tian, \
Tianfeng~Long, \
Yichi~Zhang, \
Zhengliang~Cai,\
Ruike~Chen,  \
Jifan~Zhao, Yanqing~Shen\textsuperscript{*}

\paragraph{Evaluation.} 
Ruochuan~Shi, \
Zihan~Tang, \
Jing~Lyu, \
Runze~Xiao\textsuperscript{*}

\paragraph{Real Robot.}
Jing~Lyu, \
Wenxing~Tan, \
Ningbo~Zhang,\ 
Yangtao~Hu,\
Euan~Liu,\
Yuming~Gao,\
Xiansheng~Chen, \
Junkai~Zhao,\
Runze~Xiao\textsuperscript{*}

\paragraph{Downstream Post-Training.}
Senwei~Xie, \
Huaihai~Lyu, \
Congsheng~Xu,\
Boan~Zhu,\
Ziqi~Wang

\paragraph{Infrastructure.} 
Yupu~Feng, \
Qiongqiong~Zhang

\subsection{Contributors} 

\paragraph{Infrastructure.}
Yingli~Zhao,\
Yulong~Ao

\paragraph{Real-Robot Data.}
Shaoxuan~Xie, \
You~Liu, \
Guocai~Yao

\paragraph{Product \& Operations.}
Leiduo~Zhang, \
Xiaodan~Liu, \
Yunyan~Zhang, \
Yance~Jiao

\paragraph{Brand Management.}
Xinyan~Yang,\
Jiaxing~Wei

\paragraph{Platform Management.}
Xu~Liu,\
Tengfei~Pan

\paragraph{System Management.}
Shaokai~Nie,\
Chunlei~Men

\subsection{Expert Consultant (Ordered by English last name alphabetically)}
Sen~Cui, \
Xiaojie~Jin, \
Hongyang~Li, \
Jianlan~Luo, \
Yao~Mu, \
Yunchao~Wei, \
Jun~Yan, \
Hang~Zhao, \
Xiaolong~Zheng

\subsection{Research Leads}
Jiaming~Li, \ 
Yonghua~Lin, \ 
Tiejun~Huang, \ 
Zhongyuan~Wang\textsuperscript{\Letter}, \ 
Pengwei~Wang\textsuperscript{\Letter}

\vspace{5ex}

\makeatletter
\begingroup
  \renewcommand{\thefootnote}{}
  \renewcommand{\@makefntext}[1]{\noindent #1}
  \footnotetext{%
    \small
    \textsuperscript{*} Equal Contribution.
    \par
    \textsuperscript{$\dagger$} Project Lead.\par
    \textsuperscript{\Letter} Corresponding Authors: zhongyuan@baai.ac.cn, pwwang@baai.ac.cn}
\endgroup
\makeatother

\begingroup
\linespread{1}\selectfont
\renewcommand{\refname}{}
\section*{References}\phantomsection\label{sec:references}
\setnavsection{sec:references}
\vspace{-10ex}
\bibliographystyle{BAAI-TechReport}

\bibliography{TR_Ref}
\endgroup

\newpage

\appendix

\section*{Appendix}\phantomsection\label{sec:appendix}
\setnavsection{sec:appendix}

\section{Orca Conception}\label{sec:appendix_conception}
\setnavsection{sec:appendix_conception}

\renewcommand{\thefigure}{A\arabic{figure}}
\renewcommand{\theHfigure}{A\arabic{figure}}
\setcounter{figure}{0}

\renewcommand{\thetable}{A\arabic{table}}
\renewcommand{\theHtable}{A\arabic{table}}
\setcounter{table}{0}

\begin{figure}[!htbp]
    \centering
    \includegraphics[width=1\linewidth]{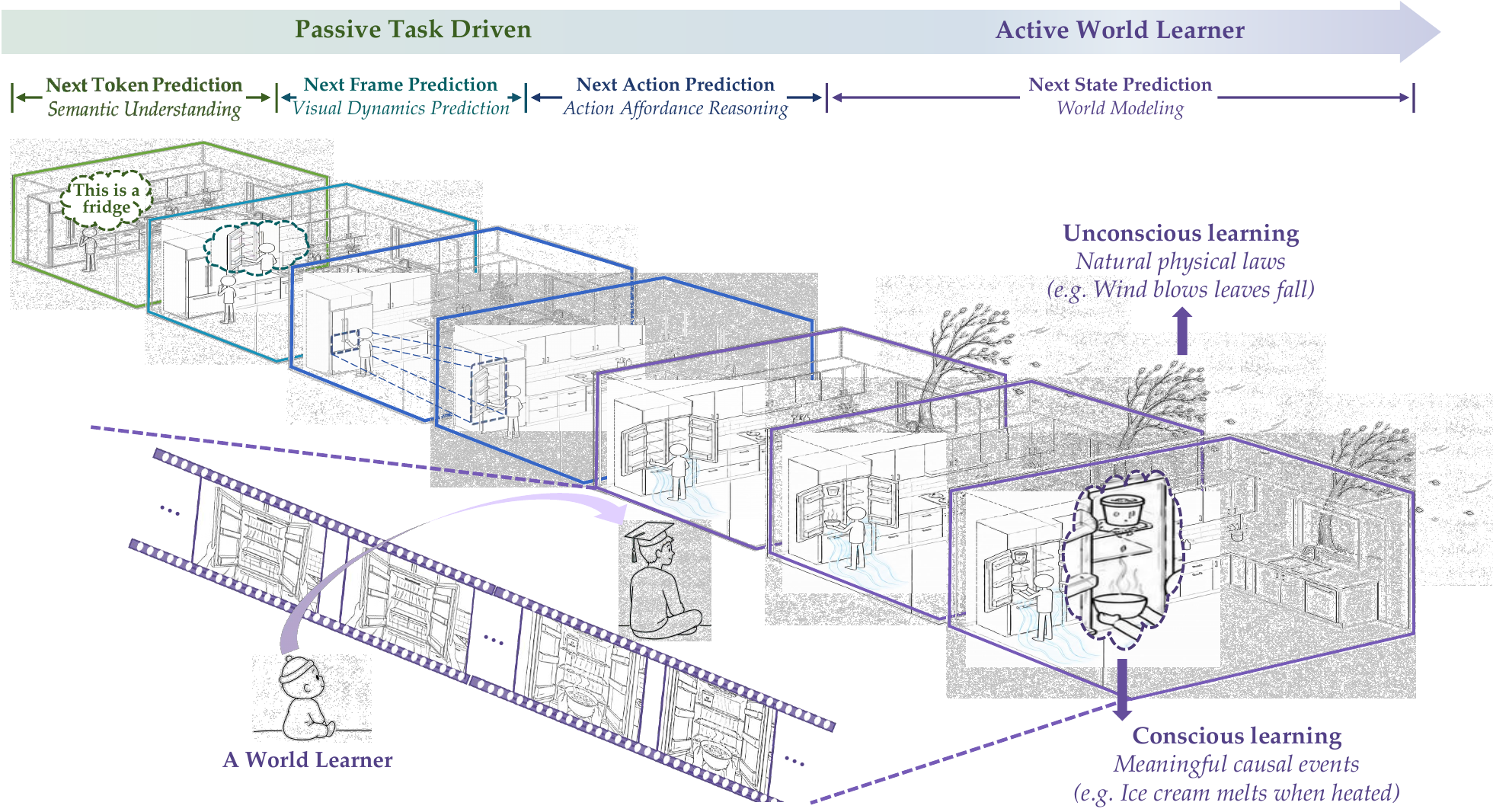}
    \caption{\textbf{Conceptual illustration of Orca}. Existing models are often organized around passive task-driven prediction, including next-token, next-frame, and next-action prediction. Orca shifts the modeling target toward next-state prediction, where multimodal world signals are used to learn a unified world latent. Unconscious learning captures dense natural dynamics from continuous observation, while conscious learning captures meaningful state transitions guided by language, events, and intentions. The learned world latent supports downstream readouts to language, vision, and action.}
    \label{fig:orca_conception}
\end{figure}

Figure~\ref{fig:orca_conception} summarizes the philosophy behind Orca. We view the development of foundation models as a transition from \textit{passive task-driven models} to an \textit{active world learner}. Existing paradigms are often centered around the output they predict: language models perform next-token prediction for semantic understanding, image and video generation models perform next-frame prediction for visual dynamics, and embodied models perform next-action prediction for action affordance. Although these paradigms produce strong task-level capabilities, their modeling targets remain tied to specific modalities.

Orca instead treats the latent state of the world as the central object of modeling. Language, vision, and action are regarded as different observations or readouts of the same underlying world state. This motivates our shift from \textit{next-token}, \textit{next-frame}, and \textit{next-action prediction} to \textbf{\textit{next state prediction}}. The goal is to learn an internal world representation that can support understanding, prediction, and intervention across different downstream interfaces.

This idea is realized through two complementary learning modes. \textbf{Unconscious learning} absorbs natural dynamics from continuous visual experience, allowing the model to learn dense state transitions and physical regularities without explicit task labels. \textbf{Conscious learning} introduces language, events, instructions, and questions as semantic conditions, allowing the model to learn meaningful state transitions associated with causal explanations and task intentions. Together, they allow Orca to internalize the world as a predictive latent representation.

Once learned, this world latent can be read out through different interfaces: to language for explanation and reasoning, to vision for prediction and imagination, and to action for intervention.

\section{Related Work}\label{sec:appendix_related_work}
\setnavsection{sec:appendix_related_work}

\renewcommand{\thefigure}{B\arabic{figure}}
\renewcommand{\theHfigure}{B\arabic{figure}}
\setcounter{figure}{0}

\renewcommand{\thetable}{B\arabic{table}}
\renewcommand{\theHtable}{B\arabic{table}}
\setcounter{table}{0}

We organize related work according to the primary learning objective at the center of each paradigm, rather than by all downstream capabilities a model may exhibit. Under this view, some unified models span multiple capability domains, but are categorized by the dominant training formulation.

\vspace{-1em}
\subsection{Self-Supervised Learning}\label{sec:ssl}

\paragraph{Latent World Models.}\label{sec:jepas}
Latent world models shift predictive learning from reconstructing high-entropy observations to modeling task-related latent. Joint Embedding Predictive Architecture (JEPA)-style work established this path by predicting semantic representations rather than pixels. I-JEPA \citep{I-JEPA}, VL-JEPA \citep{VL-JEPA}, and MC-JEPA \citep{MC-JEPA} establish joint embedding predictions for images, texts, and videos. The most influential JEPA-style works are the V-JEPA series. V-JEPA \citep{V-JEPA} demonstrates that robust video representations can be learned solely from feature predictions, without pixel reconstruction. V-JEPA 2 \citep{V-JEPA-2} combines large-scale internet video pre-training with action-based understanding, prediction, and planning. This model marks a crucial step towards embodied latent world modeling. V-JEPA 2.1 \citep{V-JEPA-2.1} enhances dense and interaction-sensitive video representations. Recent work has further advanced this paradigm in several complementary directions. LeJEPA \citep{LeJEPA} improves the stability and scalability of JEPA-style training with fewer heuristics. Causal-JEPA \citep{Causal-JEPA} introduces object-level masking and latent interventions, moving latent predictions towards object-centric causal world modeling. LeWorldModel \citep{LeWorldModel} demonstrates that lightweight end-to-end world models can be trained directly from pixels under a stable joint embedding prediction objective. GeoWorld \citep{GeoWorld} further explores the hyperbolic potential dynamics for multi-step visual planning.

\vspace{-0.5em}
\paragraph{How Orca differs.}
JEPA-style models demonstrate the effectiveness of latent prediction for self-supervised visual representation learning, which is closely related to Orca's observation-only state transition.
Orca starts from a broader world-learning formulation: it abstracts latent world states from multimodal world signals and places state transition at the center of modeling.
Specifically, Orca models state transitions under implicit dynamics and explicit semantic conditions, covering both transition directions: predicting future states and backtracking to past states.

\vspace{-1em}
\subsection{Next Token Prediction}\label{sec:ntp}
\paragraph{Large Language Models.}\label{sec:llms}
Since the development of autoregressive large language models \citep{DeepSeek-R1, Llama-2}, recent representative works have further pushed the paradigm along several directions. 
LLaMA 3.1 \citep{llama3.1} scales dense models to stronger general reasoning and instruction following. DeepSeek-V4 \citep{DeepSeek-V4} developed large-scale Mixture-of-Experts (MoE) models toward cost-effective million-token contexts. Qwen3 \citep{Qwen3} explores hybrid reasoning within a unified framework. Kimi K2 \citep{Kimi-K2} and GLM-5 \citep{GLM-5} strengthen agentic language modeling. The former emphasizes tool-oriented intelligence, and the latter focuses on long-horizon agentic engineering. MiniMax-M2.7 \citep{MiniMax-M2.7} investigates self-evolving for real-world productivity. Phi-4-reasoning \citep{Phi-4-reasoning} shows the effectiveness of high-quality reasoning supervision in dense models. These works advance ``next token prediction'' as a strong scalable approach for language intelligence.

\vspace{-0.5em}
\paragraph{Multimodal Large Language Models.}\label{sec:mllms}
In the context of multimodal environments, many works have emerged. These include instruction-tuned visual language models, native multimodal foundation models, and agents.
LLaVA \citep{llava} pioneered a widely adopted approach to connect vision encoders to large language models. It leverages multimodal instruction tuning to achieve general visual language understanding. Gemini 3.1 \citep{Gemini-3.1-pro} enhanced multimodal reasoning, agent usage, and long context processing capabilities. The latest Qwen series \citep{Qwen3VL, qwen36_35b_a3b} further strengthens visual perception and reasoning, spatial and video dynamic understanding capabilities, and formally moves towards native multimodal large models. GPT-5.4 \citep{GPT54} extends cutting-edge multimodal systems to specialized knowledge work, native computer applications, and agent usage. Gemma 4 \citep{Gemma4} emphasizes cutting-edge multimodal intelligence on the device, supporting image, text, and audio input as well as long context processing. Kimi K2.5 \citep{KimiK2.5} developed multimodal models toward visual agentic intelligence by jointly optimizing text and vision, and introducing Agent Swarm, which decomposes complex tasks into heterogeneous sub-problems for concurrent execution. LLaMA 4 \citep{llama4} introduces a hybrid expert and native multimodal generative model, while Mistral Medium 3.5 \citep{Mistral-Medium-3.5} integrates instruction tracking and inference, and is equipped with a visual encoder trained for different image sizes and aspect ratios. The MLLM paradigm has also been extended toward embodied reasoning and robotic manipulation. RoboBrain series \citep{RoboBrain, RoboBrain2.0,reasonrft} proposes an MLLM-based robotic brain model that integrates general multimodal data with robot-specific supervision. These models improve multimodal understanding and agentic interaction capabilities.

The unified multimodal models also begin to blur the boundary between token prediction and frame prediction. For example, Emu3 \citep{Emu3} and Emu3.5 \citep{Emu35} unify multimodality under the ``next-token-prediction'' paradigm. Emu3.5 significantly optimizes the slow image generation speed and poor performance of this autoregressive paradigm. BAGEL \citep{BAGEL} follows the next-token-prediction paradigm and uses a MoT \citep{MoT} architecture. It exhibits emergent capabilities such as image generation, image editing, and future frame prediction. Janus-Pro \citep{Janus-Pro} deploys a unified autoregressive Transformer architecture and decouples the visual encoding path for multimodal understanding and generation. Cosmos 3 \citep{Cosmos3} is an omnimodal world model for Physical AI. It jointly processes and generates language, image, video, audio, and action sequences within a unified MoT architecture.

\vspace{-0.5em}
\paragraph{How Orca differs.}
Next-token models organize knowledge and reasoning through autoregressive language modeling.
Orca uses language as an explicit semantic condition for state transition: language can specify events, task intentions, and causal premises that guide how the current state transitions toward a target state.
Meanwhile, VQA response generation preserves the language interface and strengthens the commonsense and semantic grounding of the learned world representation.

\vspace{-1em}
\subsection{Next Frame Prediction}\label{sec:nfp}
\vspace{-0.5em}
\paragraph{Image Generation Models.}\label{sec:vgms}
From a macro perspective, the representative image generation models can be viewed as frame-level prediction models. These models map language or multimodal conditions to target visual observations, thus extending prediction from the token space to the visual frame space. Nano Banana Pro  \citep{nanobananapro} boasts powerful multilingual text rendering, infographic creation, and search-based visualization capabilities. GPT Image 2 \citep{ChatGPT-Images-2.0} extends the image generation product line to native text and image model interfaces, supporting high-quality image generation and editing. Qwen-Image \citep{Qwen-image} uses the MMDiT \citep{SD3} to achieve complex text rendering and precise image editing. FLUX.2 \citep{flux2} drives the development of professional-grade image generation through high editing consistency, powerful cue tracking, multi-reference control, and real-time network context-based generation. Overall, these models significantly advance the next frame towards higher fidelity, instruction tracking, and image editing.

\vspace{-0.5em}
\paragraph{Video Generation Models.}\label{sec:videogms}
Video generation works have expanded frame-level prediction from static visual observation to temporally coherent visual sequences~\citep{videoworld,VideoWorld-2,cognitive_map}. Seedance 2.0 \citep{seedance2.0} supports text, image, audio, and video inputs and possesses powerful reference-based generation and editing capabilities. Sora advances text-to-video generation to longer, higher-quality videos and attempts to understand and simulate the physical world of motion. Cosmos-Predict 2.5 \citep{cosmospredict2.5} unifies ``Any2World'' into a single framework, aiming to support pixel-level world modeling. Wan2.1 \citep{wan2.1} provides a comprehensive set of diffusion transform-based video foundational models. Overall, these models significantly improve temporal coherence, controllability, and multimodal conditional effectiveness.

\vspace{-0.5em}
\paragraph{How Orca differs.}
Image and video generation models are often regarded as world models because they can synthesize coherent and visually appealing frames.
Orca's goal is not to create a painter, but to model whether a target state follows the physical constraints and interaction process of the real world.
Therefore, Orca emphasizes action execution, scene consistency, physical plausibility, scene and object consistency, contact relationships, and instruction following in interaction-conditioned state transitions.

\vspace{-1em}
\subsection{Next Action Prediction}\label{sec:nap}
\paragraph{Vision Language Action Models.}\label{sec:vlas}
Vision Language Action (VLA) models, an architecture in embodied intelligence, provide a feasible path to improve generalization and multi-task learning capabilities~\citep{manipulation_survey, bai2026latentreasoningvla,lyu2026general,lyu2026last,liu2026pi_0}. OpenVLA \citep{OpenVLA}, which first presented an open-source VLA model built on a pre-trained VLM, showing strong cross-embodied general manipulation capabilities after training on large-scale robot data. 
$\pi_{0.5}$ \citep{pi0.5} builds upon $\pi_{0}$ \citep{pi0} by employing a collaborative training method based on heterogeneous data, including multiple robots, subtasks, and network data. It improves generalization capabilities in open worlds. $\pi_{0.7}$ \citep{pi0.7} uses task descriptions, generated sub-target images, and episode metadata to advance the general robot foundation model further. GR00T (N1 \citep{GR00TN1} to N1.7 \citep{GR00TN1.7}) gives a foundation model for general-purpose humanoid robots, achieving better performance in dual-arm and mobile manipulation tasks. VLA-Adapter \citep{VLA-Adapter} explores which information in VLM is more conducive to action generation, and its convergence scheme achieves better performance on a tiny backbone than large-scale VLAs. SimpleVLA-RL \citep{SimpleVLA-RL} develops an RL framework for VLA, improving long-horizon planning and generalization capabilities under limited demonstration, and surpassing supervised fine-tuning models \citep{han2026dexhil}. VLA-RFT \citep{VLA-RFT} is the first work to use a world model for post-training a VLA model. It treats the world model as a simulator and provides dense validation rewards.

\vspace{-0.5em}
\paragraph{Video Action Models.}\label{sec:vams}
Some researchers argue that using a static VLM as the backbone increases the learning burden of the action expert because it needs to model both visual dynamics and control information simultaneously. To alleviate this problem, some works attempt to introduce visual dynamics modeling capabilities from video generation models to reduce the learning difficulty of the action expert. VPP \citep{VPP} is an early representative work in this direction, learning general robot policies by combining implicit inverse dynamics models with predictive visual representations extracted from video models. UVA \citep{UVA} achieves efficient action inference by learning a shared video-action latent with a decoupled diffusion head and combining forward/backward dynamics to jointly optimize observations and action prediction. Mimic-video \citep{mimic-video} combines a video model and an action decoder, demonstrating strong generalization ability and sample efficiency. Cosmos-Policy \citep{cosmos-policy} adapts the pre-trained Cosmos-Predict \citep{cosmospredict2.5} into a robot policy through single-stage post-training. This model predicts future states and value functions for planning.

\vspace{-0.5em}
\paragraph{World Action Models.}\label{sec:wams}
With the recent surge in interest in world models, some embodied models have begun to jointly model action and future world states. Motus \citep{motus}, based on a MoT architecture, constructs a unified latent action world model, fusing understanding, video generation, and action experts. DreamZero \citep{DreamZero} constructs a world action model as a zero-shot policy based on a pre-trained video model, achieving real-time closed-loop manipulation by predicting future observations and actions. GigaWorld-Policy \citep{GigaWorld-Policy} proposes an efficient, action-centric world action model. This model decouples action prediction from video generation, improving inference efficiency while preserving visual dynamic supervision. Being-H0.7 \citep{Being-H0.7} proposes a latent world-action model that connects action prediction and world modeling through a compact latent inference space, injecting future-aware inference into action generation. VLA-JEPA \citep{VLA-JEPA} and JEPA-VLA \citep{JEPA-VLA} link latent predictive embeddings with VLA and imitation learning.

\vspace{-0.5em}
\paragraph{How Orca differs.}
VLA and world-action models usually organize embodied learning around action prediction, policy learning, or joint video-action modeling.
Orca follows a world-learning-first philosophy: it first learns how scenes and objects change through state-transition modeling, without relying on action labels during pre-training.
By building a stronger world representation of temporal changes, object motion, and local physical interactions, Orca provides a more general foundation that can adapt more efficiently to embodied tasks under limited robot data and OOD settings.
\section{Training Settings}
\setnavsection{sec:appendix_training}
\label{sec:appendix_training}

\renewcommand{\thefigure}{C\arabic{figure}}
\renewcommand{\theHfigure}{C\arabic{figure}}
\setcounter{figure}{0}

\renewcommand{\thetable}{C\arabic{table}}
\renewcommand{\theHtable}{C\arabic{table}}
\setcounter{table}{0}

\renewcommand{\theequation}{C-\arabic{equation}}
\renewcommand{\theHequation}{C-\arabic{equation}}
\setcounter{equation}{0}

This appendix provides implementation details for the training procedure described in \textbf{Section~\ref{sec:training}}.

\subsection{Pre-Training Settings}
\label{sec:appendix_pretraining_settings}

\subsubsection{Pre-Training Objective Definitions}
\label{sec:appendix_pretraining_objective}
This section specifies the loss definitions used for Orca pre-training. The pre-training objective contains three terms: \textit{1) observation-only state transition} \(\mathcal{L}_{\mathrm{obs}}\), \textit{2) event-conditioned state transition} \(\mathcal{L}_{\mathrm{evt}}\), and \textit{3) VQA response generation} \(\mathcal{L}_{\mathrm{vqa}}\). 
\textit{1) observation-only state transition} and \textit{2) event-conditioned state transition} are supervised by visual latent extracted by the frozen vision encoder from the VLM backbone, and \textit{3) VQA response generation} is supervised with the standard next-token prediction. For the first two state transitions, Orca matches the predicted latent to the ground truth latent extracted by the frozen vision encoder. \(\hat{v}^l\) and \(v^l\) denote the predicted and ground truth latent at the visual-token positions. We use the latent matching: 

\vspace{-2ex}
\begin{equation}\label{eq:appendix_latent_matching} 
\ell_{\mathrm{lat}}(\hat{v}^l,v^l) = 0.1\,\|\hat{v}^l-v^l\|_2^2 + 0.9 \left( 1- \frac{\langle \hat{v}^l, v^l\rangle} {\|\hat{v}^l\|_2\|v^l\|_2} \right). 
\end{equation}

\begin{enumerate}[
    leftmargin=1.35em,
    itemsep=0.35em,
    topsep=0.35em,
    parsep=0em
]
    \item[] \textit{\textbf{1) Observation-only state transition.}} \(v^l_{t+1}\) is the latent of the next frame. The loss is:
\end{enumerate}

\begin{equation} 
\mathcal{L}_{\mathrm{obs}} = \mathbb{E} \left[ \ell_{\mathrm{lat}} \left( \hat{v}^l_{t+1}, v^l_{t+1} \right) \right]. \label{eq:appendix_obs_loss} 
\end{equation}

\vspace{-2ex}
\begin{enumerate}[
    leftmargin=1.35em,
    itemsep=0.35em,
    topsep=0.35em,
    parsep=0em
]
    \item[] \textit{\textbf{2) Event-conditioned state transition.}} The language specifies whether the current state should be mapped toward an adjacent (earlier or later) event state. Accordingly, Orca predicts the visual latent in the previous event selected by the \textit{previous-event condition} and the visual latent in the next event selected by the \textit{next-event condition}. The event-conditioned loss averages the latent-matching from the two transition directions:
\end{enumerate}

\vspace{-2ex}
\begin{equation} 
\mathcal{L}_{\mathrm{evt}} = \frac{1}{2} \mathbb{E} \left[ \ell_{\mathrm{lat}} \left( \hat{v}^l_{\mathrm{prev}}, v^l_{\mathrm{prev}} \right) + \ell_{\mathrm{lat}} \left( \hat{v}^l_{\mathrm{next}}, v^l_{\mathrm{next}} \right) \right]. 
\label{eq:appendix_event_loss}
\end{equation} 

\begin{enumerate}[
    leftmargin=1.35em,
    itemsep=0.35em,
    topsep=0.35em,
    parsep=0em
]
    \item[] \textit{\textbf{3) VQA response generation.}} Orca uses the language modeling head to predict the target answer with the standard next-token prediction loss. This term is denoted as \(\mathcal{L}_{\mathrm{vqa}}\).
\end{enumerate}

The final Orca's pre-training objective is: $\mathcal{L}_{\mathrm{pre}} = 0.1\,\mathcal{L}_{\mathrm{obs}} + 0.5\,\mathcal{L}_{\mathrm{evt}} + 0.4\,\mathcal{L}_{\mathrm{vqa}}$. At the data-sampling level, Orca mixes state transition samples and VQA samples with an approximate ratio of \(5:1\).

\subsubsection{Query-Based Implementation} \label{sec:appendix_pretraining_query_implementation} 
\begin{figure}[!htbp]
    \centering
    \includegraphics[width=1\linewidth]{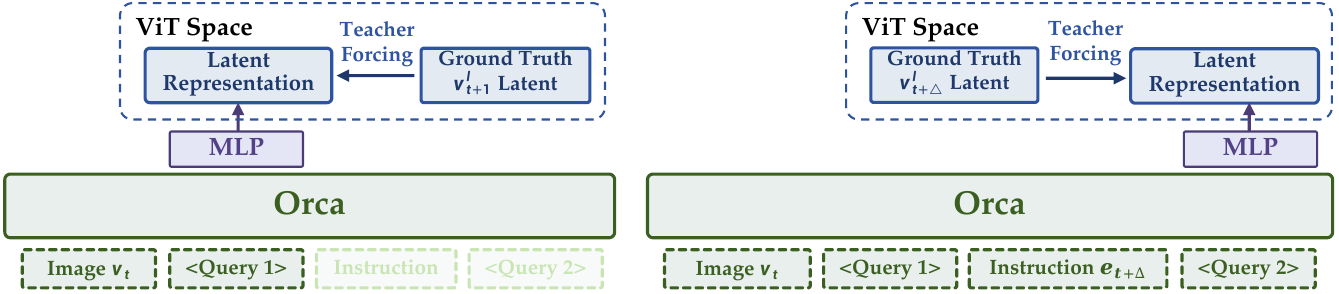}
    \caption{\textbf{The implementation of Queries.}}
    \label{fig:query}
\end{figure}

We provide the implementation details of the query-based training described in \textbf{Section~\ref{sec:pretraining_recipe}}. \textit{Note that all queries are trained from scratch}. The implementation of queries is shown in Figure \ref{fig:query}.

\begin{enumerate}[
    leftmargin=1.35em,
    itemsep=0.35em,
    topsep=0.35em,
    parsep=0em
]
    \item[] \textit{\textbf{1) Observation-only state transition.}} Given the current observation \(v_t\) and \textit{<Query 1>} \(q_1\), Orca predicts the latent \(\hat{v}^l_{t+1}\) of a temporally next frame. The last-layer hidden state of \(q_1\) is passed to the \textit{visual transition head} (two-layer MLP), and the ground truth latent \(v^l_{t+1}\) is obtained by the frozen vision encoder of VLM backbone. 
    \item[] \textit{\textbf{2) Event-conditioned State Transition.}} Given \(v_t\), \(q_1\), an instruction \(e_{t+\Delta}\), and the \textit{<Query 2>} \(q_2\), Orca predicts the latent \(\hat{v}^l_{t+\Delta}\) of random frame in the instruction-specified target event. \(e_{t+\Delta}\) specifies the transition direction and target event, while \(q_2\) reads out the corresponding instruction-conditioned predictive state. The previous-event \(\mathcal{L}_{\mathrm{prev}}\) and next-event directions \(\mathcal{L}_{\mathrm{next}}\), which are calculated \(\mathcal{L}_{\mathrm{evt}}\) in Equation~\ref{eq:appendix_event_loss}. 
\end{enumerate}

\subsubsection{Pre-Training Hyperparameters} \label{sec:appendix_pretraining_hparams} 
Table~\ref{tab:appendix_pretraining_hparams} reports the main pre-training hyperparameters. The table groups the model-scale settings, optimization settings, and objective-specific settings.

\begin{table}[!htbp]
\centering
\caption{\textbf{Orca pre-training hyperparameters.}} 
\label{tab:appendix_pretraining_hparams} 
\resizebox{\linewidth}{!}{%
\begin{tabular}{lcc} 
\toprule[1pt] 
\textbf{Hyperparameter} & \textbf{Orca-4B} & \textbf{Orca-0.8B} \\ 
\midrule 
    Base VLM & Qwen3.5-4B & Qwen3.5-0.8B \\ 
    Backbone hidden size \(H\) & 2560 & 1024 \\ 
    Training resources & 32 nodes / 256 GPUs & 32 nodes / 256 GPUs \\
    State-transition per-GPU batch size & 8 & 8 \\ 
    State-transition gradient accumulation & 2 & 2 \\
    VQA per-GPU batch size & 4 & 4 \\
    Training steps & 10,844 & 10,844 \\ 
    Approximate video hours & 12.5K h & 12.5K h \\ 
    Maximum sequence length & 1024 & 1024 \\ 

    \midrule 
    Optimizer & AdamW & AdamW \\
    Base VLM learning rate & \(3.5\times 10^{-5}\) & \(3.5\times 10^{-5}\) \\ 
    Visual head & 
    2$\times$MLP, \(2560\!\rightarrow\!20480\!\rightarrow\!2560\) &
    2$\times$MLP, \(1024\!\rightarrow\!8192\!\rightarrow\!1024\) \\
    Visual head learning rate & \(1.2\times 10^{-4}\) & \(1.2\times 10^{-4}\) \\ 
    Visual encoder / ViT & Frozen & Frozen \\
    LLM & Trainable & Trainable \\
    Adam betas & \([0.9, 0.95]\) & \([0.9, 0.95]\) \\ 
    Weight decay & \(1\times 10^{-8}\) & \(1\times 10^{-8}\) \\ 
    Scheduler & Cosine with minimum LR & Cosine with minimum LR \\ 
    Warmup steps & 200 & 200 \\ 
    Minimum learning rate & \(1\times 10^{-6}\) & \(1\times 10^{-6}\) \\
    
    \midrule 
    Latent matching loss & 
    \(0.1\,\mathrm{MSE}+0.9\,\mathrm{Cosine}\) &
    \(0.1\,\mathrm{MSE}+0.9\,\mathrm{Cosine}\) \\
    Observation-only coefficient \(\mathcal{L}_{\mathrm{obs}}\) & 0.1 & 0.1 \\ 
    Event-conditioned coefficient \(\mathcal{L}_{\mathrm{evt}}\) & 0.5 & 0.5 \\
    VQA coefficient \(\mathcal{L}_{\mathrm{vqa}}\) & 0.4 & 0.4 \\
    Number of queries & 256 & 256 \\
    \bottomrule[1pt] 
\end{tabular}%
}
\end{table}

\subsection{Downstream Readout Post-Training Settings}
\label{sec:appendix_downstream_readout_settings}

This section provides implementation details for the downstream readout training described in \textbf{Section~\ref{sec:downstream_readout}}. The overall readout architecture is illustrated in Figure~\ref{fig:downstream_readout_architecture}. Here, we further detail the implementation of the language readout, the SD3.5-based vision readout, and the DiT-based action readout.

\subsubsection{Language Readout}
\label{sec:appendix_language_readout}

The language readout does not introduce an additional trainable module.
It reuses the language modeling head of the VLM backbone as its interface. Given a visual observation and an instruction, Orca produces the response autoregressively with the LM head. This readout exposes the learned latent through natural language and is used for VQA, event-level interpretation, and causal explanation.

\subsubsection{Vision Readout}
\label{sec:appendix_vision_readout}

The vision readout maps Orca's predicted visual latent state into image space. We instantiate this readout with a pretrained \textit{Stable Diffusion 3.5} decoder~\citep{sd35}. During readout training, the \textit{VAE} and \textit{MMDiT} weights of SD3.5 are kept frozen, while the \textit{MLP adaptor} and \textit{LoRA}~\citep{lora} attached to the decoder attention projections are trainable. The learned latent is fed into one path of \textit{MMDiT} after passing through an \textit{MLP adaptor}; the target image is denoised, fed into a frozen \textit{VAE}, and then fed into another path of \textit{MMDiT}. Finally, the predicted image is obtained through a multi-step denoise.

\paragraph{Settings.}
The architecture of the \textit{MLP adaptor} and the main training settings of the \textit{SD3.5} vision readout are summarized in Table~\ref{tab:appendix_vision_readout_settings}. The \textit{MLP adaptor} projects Orca's visual-state latent into the \textit{SD3.5} joint conditioning space, including token-level conditions and an auxiliary pooled condition. During decoder training, the target image is resized to \(768\times768\).

\begin{table}[!htbp] 
\centering 
\caption{\textbf{The vision readout settings.}} \label{tab:appendix_vision_readout_settings} 
\begin{tabular}{ll} 
\toprule[1pt]
\textbf{Item} & \textbf{Settings} \\ 
\midrule
\rowcolor[rgb]{ .949, .949, .949} \multicolumn{2}{l}{\textit{Architecture}} \\
Base image decoder & Stable Diffusion 3.5 MMDiT \\ 
Condition input & Orca's latent, \((64, 2560)\) \\ 
Input projection & LayerNorm (2560), Linear (2560 \(\rightarrow\) 4096) \\
Residual MLP blocks & 4 blocks \\ 
Block width & 4096 \(\rightarrow\) 16384 \(\rightarrow\) 4096 \\ 
Output normalization & LayerNorm (4096) \\ 
Pooled branch & Mean pooling and two-layer MLP to 2048 \\ 
LoRA target modules & Attention projections \\ 
Trainable adaptor parameters & 556.9M \\ 
\midrule \rowcolor[rgb]{ .949, .949, .949} \multicolumn{2}{l}{\textit{Training}} \\ 
Frozen modules & SD3.5 VAE and MMDiT weights \\ 
Trainable modules & MLP adaptor and LoRA parameters \\ 
Target image size & \(768 \times 768\) \\ 
Global batch size & 512 \\ 
Training steps & 200,000 \\ 
Optimizer & AdamW \\ 
Adaptor learning rate & \(1 \times 10^{-4}\) \\ 
LoRA learning rate & \(5 \times 10^{-5}\) \\ 
Weight decay & 0.01 \\ 
Scheduler & OneCycleLR with cosine annealing \\ 
LoRA rank / alpha / dropout & 32 / 32 / 0.05 \\ 
\bottomrule[1pt] 
\end{tabular} 
\end{table}

\subsubsection{Action Readout}
\label{sec:appendix_action_readout}
The action readout maps Orca's latent to action chunks to control robot manipulation. It takes Orca's learned \textit{latent}, \textit{noisy action} with time embedding, and robot \textit{proprioception} as inputs. A DiT-based~\citep{dit} Action Expert with flow-matching~\citep{flow_matching} loss then predicts a short-horizon action chunk. The action expert uses the following conditions: 

\begin{enumerate}[
    leftmargin=1.35em,
    itemsep=0.35em,
    topsep=0.35em,
    parsep=0em
]
    \item[] \textbf{\textit{1) Latent \(q_1\)}}: predictive query states from Orca, providing latent for future state evolution. 
    \item[] \textbf{\textit{2) Noisy action with time embedding}}: Actions with Gaussian noise, and time embedding added.
    \item[] \textbf{\textit{3) Proprioception}}: robot proprioceptive state, including joint and end-effector related information. 
\end{enumerate}

\paragraph{Settings.} 
The \textit{Action Expert} is trained with the flow-matching loss to obtain the action chunks. The ground-truth action chunk is perturbed with Gaussian noise, and the \textit{Action Expert} predicts the corresponding velocity.
The architecture and training settings of the \textit{Action Expert} are shown in Table~\ref{tab:appendix_action_readout_settings}.

\begin{table}[!htbp] 
\centering 
\caption{\textbf{The action readout settings.}} \label{tab:appendix_action_readout_settings} 
\begin{tabular}{lc} 
\toprule[1pt]
\textbf{Item} & \textbf{Settings} \\ 
\midrule
\rowcolor[rgb]{ .949,  .949,  .949}\multicolumn{2}{l}{\textit{Architecture}} \\
Action expert type & DiT-based model with flow-matching loss \\ 
DiT blocks & 8 \\ 
Block pattern & Interleaved a self attention and a cross attention \\
Conditions & \(q_{1}\), Noisy action, Proprioception \\ 
Input embedding dimension & 768 \\ 
Hidden size & 1024 \\ 
Attention heads & 12 \\ 
Action dimension & 16 \\ 
State dimension & 16 \\ 
Action horizon & 30 \\ 
Repeated samples & 8 \\ 
Inference timesteps & 4 \\ 
Position embedding & Enabled \\ 
\midrule 
\rowcolor[rgb]{ .949,  .949,  .949}\multicolumn{2}{l}{\textit{Training}} \\
Global batch size & 128 \\ 
Training steps & 20,000 \\ 
AMP (Automatic Mixed Precision) & True \\ 
Gradient clipping norm & 1.0 \\ 
Optimizer & AdamW \\ 
Action expert learning rate & \(1 \times 10^{-4}\) \\ 
Orca backbone & Frozen \\ 
Adam betas & \([0.9, 0.95]\) \\
Weight decay & \(1 \times 10^{-8}\) \\ Scheduler & Cosine with minimum LR \\ Warmup steps & 500 \\ Minimum learning rate & \(1 \times 10^{-6}\) \\ 
\bottomrule[1pt]
\end{tabular}
\end{table}
\section{Infrastructure}\label{sec:appendix_infrastructure}
\setnavsection{sec:appendix_infrastructure}

\renewcommand{\thefigure}{D\arabic{figure}}
\renewcommand{\theHfigure}{D\arabic{figure}}
\setcounter{figure}{0}

\renewcommand{\thetable}{D\arabic{table}}
\renewcommand{\theHtable}{D\arabic{table}}
\setcounter{table}{0}

Orca training integrates visual embedding, language modeling, future visual-latent prediction, and action-related branches, resulting in higher memory pressure and communication cost than standard VLM training. To support stable and scalable large-scale training, we build the Orca training infrastructure on FlagScale~\citep{flagscale} and restructure the training pipeline at the system level. The optimization focuses on three aspects: distributed sharding, memory-efficient execution, and communication-computation overlap. Concretely, we adapt FSDP2 (Fully Sharded Data Parallel), activation recomputation, chunked cross-entropy loss, and forward/backward pre-fetching.

\paragraph{FSDP2.} 
We migrate the training backend from DeepSpeed to FSDP2. FSDP2 supports flexible sharding of parameters, gradients, and optimizer states under different memory budgets and model scales, which improves multi-GPU training efficiency while maintaining training stability. We further enable resharding to release redundant parameter copies after use, reducing peak GPU memory during distributed training. For lightweight visual blocks, we remove unnecessary FSDP sharding to avoid excessive communication and scheduling overhead.

\paragraph{Activation Recompute.} 
We apply activation recomputation to reduce the activation memory footprint. Instead of retaining all intermediate activations, the training process checkpoints selected activation boundaries and reconstructs the required intermediate states during backpropagation. This trades moderate additional computation for substantial memory savings, enabling larger batch sizes and improving overall throughput under memory-constrained settings.

\paragraph{Chunked Cross-Entropy Loss.} 
Memory profiling shows that the cross-entropy computation in the VLM forward stage introduces a significant peak-memory spike under long-sequence and large-vocabulary settings, mainly due to the log-softmax intermediate tensor. We therefore adopt chunked cross-entropy loss, which partitions the token dimension and computes the loss block by block. This avoids materializing the full logits tensor during loss computation and reduces peak memory, leaving more memory budget for larger batch sizes and longer sequences.

\paragraph{Forward/Backward Pre-fetching.} 
Performance analysis shows that FSDP2 all-gather operations can expose communication stalls when parameter aggregation and layer computation are not sufficiently overlapped. We introduce forward/backward pre-fetching to overlap the parameter all-gather for upcoming layers with computation in the current layer. This scheduling reduces GPU idle time caused by communication waits and improves overall device utilization.

\begin{table}[!htbp]
    \centering
    \caption{\textbf{Training throughput of infrastructure optimizations on H100 GPUs.}}
    \label{tab:appendix_infra_comparison}
    \begin{tabular}{lc}
    \toprule[1pt]
    \textbf{Infrastructure} & \textbf{Samples/Sec/GPU $\uparrow$} \\
    \midrule[0.5pt]
    StarVLA~\citep{StarVLA} & 0.66 \\
    FSDP2 baseline & 0.97 \\
    \midrule[0.5pt]
    + Chunked Cross-Entropy Loss & 1.35 \\
    + Activation Recompute & 2.86 \\
    + Forward/Backward Pre-fetching (\textbf{Full Orca}) & 2.91 \\
    \bottomrule[1pt]
    \end{tabular}
\end{table}

As shown in Table~\ref{tab:appendix_infra_comparison}, the optimized infrastructure achieves 2.91 samples/sec/GPU on H100 GPUs. This corresponds to a 3.0$\times$ improvement over the FSDP2 baseline of 0.97 samples/sec/GPU and a 4.4$\times$ improvement over the StarVLA~\citep{StarVLA} training pipeline.
\section{Evaluation Settings}\label{sec:appendix_evaluation}
\setnavsection{sec:appendix_evaluation}

\renewcommand{\thefigure}{E\arabic{figure}}
\renewcommand{\theHfigure}{E\arabic{figure}}
\setcounter{figure}{0}

\renewcommand{\thetable}{E\arabic{table}}
\renewcommand{\theHtable}{E\arabic{table}}
\setcounter{table}{0}

\renewcommand{\thelstlisting}{E\arabic{lstlisting}}
\renewcommand{\theHlstlisting}{E\arabic{lstlisting}}
\setcounter{lstlisting}{0}

\subsection{Text Generation}\label{sec:appendix_evaluation_toL}

\paragraph{Benchmarks.} In terms of text generation, we use the following benchmarks for evaluation:

\begin{itemize}
    \item \textbf{MVBench} \citep{MVBench} evaluates a model's general video understanding capabilities through multiple-choice QA tasks, covering action recognition, temporal reasoning, object interaction, and event-level understanding.
    \item \textbf{TemporalBench} \citep{cai2024temporalbench} evaluates a model's ability to handle fine-grained temporal dynamics, including action frequency, motion amplitude, and event order. The test uses short video (0-20 seconds) QA tasks and employs Multiple Binary Accuracy (MBA) as the evaluation metric.
    \item \textbf{SWITCH} \citep{switch2025} evaluates a model's ability to interact with real-world TCI (Tangible Control Interface). It requires the model to possess common sense and physical understanding, causal prediction capabilities, and the ability to predict and verify operational results in the spatiotemporal dimensions.
    \item \textbf{3DSRBench} \citep{ma20253dsrbench} evaluates a model's 3D spatial reasoning capabilities, including reasoning about height, location, orientation, and multiple objects.
\end{itemize}

\paragraph{Baselines.} In terms of text generation, we use the following baselines for evaluation:

\begin{itemize}
    \item \textbf{V-JEPA 2.1} \citep{V-JEPA-2.1} is a self-supervised video model that learns dense visual representations from images and videos. It serves as a representative latent world-model baseline, emphasizing spatially grounded and temporally consistent visual understanding.
    \item \textbf{Emu3} \citep{Emu3} tokenizes images, text, and videos into a unified discrete token space and trains a single Transformer with next-token prediction. It provides a native multimodal baseline for both perception and generation.
    \item \textbf{Emu3.5} \citep{Emu35} extends the next-token prediction paradigm toward native multimodal world modeling. It learns from interleaved vision-language sequences and supports long-horizon multimodal generation and spatiotemporally consistent world exploration.
    \item \textbf{Qwen3.5} \citep{qwen35} is a native multimodal foundation model with early vision-language fusion. It provides a strong general-purpose VLM baseline for visual understanding, reasoning, long-context modeling, and agentic interaction.
    \item \textbf{Gemma 4} \citep{Gemma4} is an efficient open multimodal model family supporting text, vision, video, and audio understanding. We use it as a compact yet strong baseline for multimodal reasoning and instruction following.
    \item \textbf{DeepSeek-VL2} \citep{DeepSeek-VL2} is a Mixture-of-Experts vision-language model that improves high-resolution visual understanding through dynamic tiling and an efficient MoE language backbone. It is included as a strong VLM baseline for visual question answering, OCR, document understanding, and visual grounding.
    \item \textbf{MiniCPM-V-4.6} \citep{MiniCPM-o-4.5} is an edge-deployment-friendly multimodal model designed for efficient image and video understanding. It introduces compact visual encoding and mixed visual token compression, making it a lightweight baseline for mobile and resource-constrained scenarios.
    \item \textbf{SmolVLM2} \citep{SmolVLM} is a compact vision-language model series designed for resource-efficient image and video understanding. It is included as a lightweight baseline to evaluate whether small VLMs can capture temporal and spatial dynamics under limited model capacity.
\end{itemize}

\subsection{Image Prediction}\label{sec:appendix_evaluation_toV}
\subsubsection{Benchmarks} \label{sec:appendix_evaluation_toV_benchmark}
To evaluate the model's ability to predict state changes in real-world interactive scenarios, we developed \textbf{PRICE-V0.1} (i.e., Prediction of Real-world Interactions with Constraints Evaluation). This benchmark is an instruction-conditional image-to-image generation task (TI2I). Given an initial state image and an instruction, PRICE requires generating a target state image after the corresponding action is performed. Unlike traditional image editing tasks, PRICE focuses more on whether the model can understand the actual impact of the instruction in the real physical environment and generate state change results that conform to scene constraints and common sense.

\begin{figure}[!htbp]
    \centering
    \includegraphics[width=0.98\linewidth]{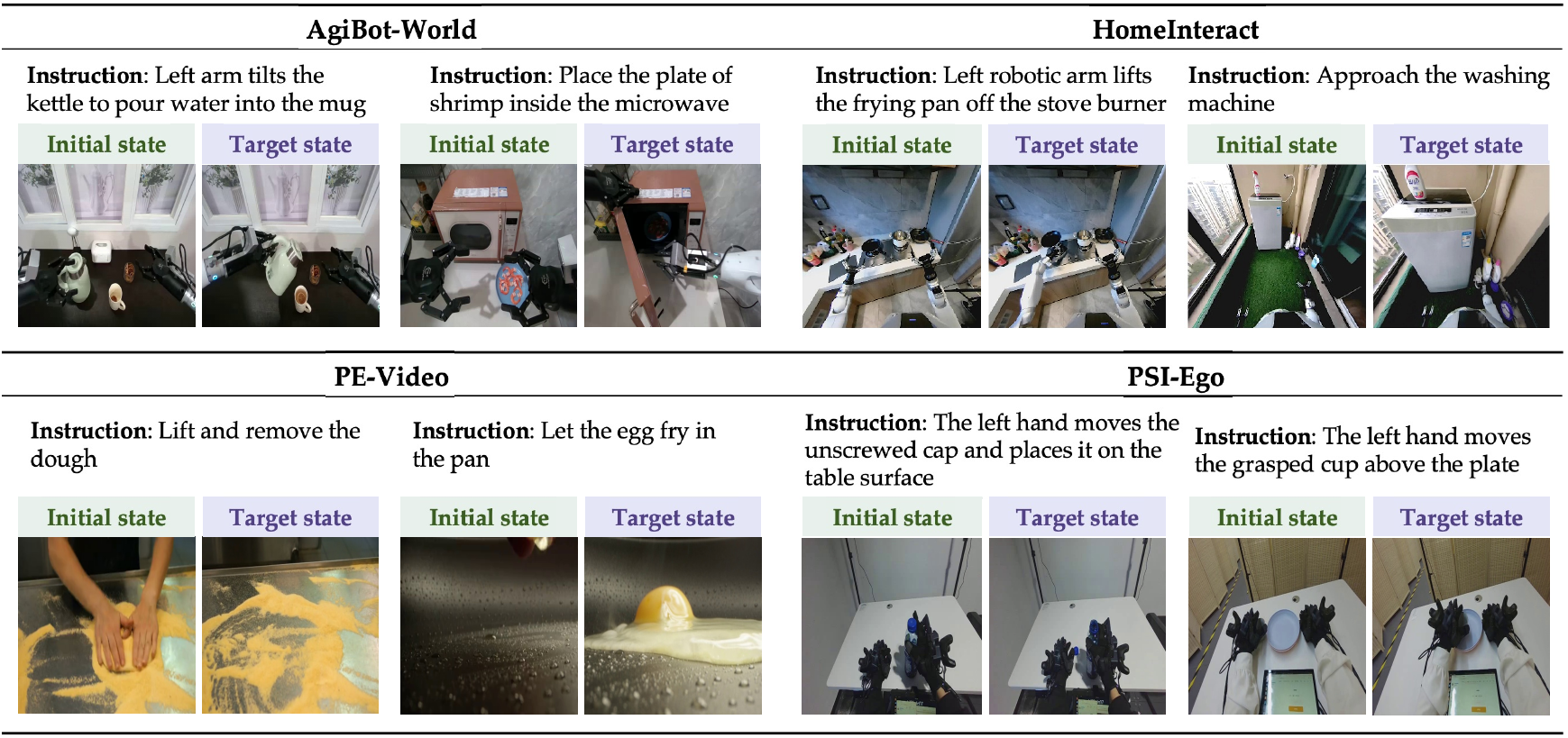}
    \caption{\textbf{PRICE-V0.1 Examples.}} %
    \label{fig:PRICE}
\end{figure}

PRICE is derived from four real-world robot or first-person perspective interaction datasets: \textit{AgiBot-World} \citep{Agibot-world}, \textit{HomeInteract}, \textit{PE-Video} \citep{PE}, and \textit{PSI-Ego} \citep{SynData}. HomeInteract is the closed-source general data collected by the dual-arm wheeled robot in the home scene. Each sample consists of three parts: an instruction, an initial state image, and a target state image, as shown in Figure \ref{fig:PRICE}. These samples cover a variety of embodied interaction scenarios, including object manipulation, changes in scene state, action-result prediction, and interaction relationships between humans and objects, or between robots and objects.

\subsubsection{Metrics} \label{sec:appendix_evaluation_toV_metrics}
We use closed-source Gemini 3.1 Pro \citep{Gemini-3.1-pro}, GPT 5.4 \citep{GPT54}, and Doubao Seed 2.0 Pro \cite{doubao-seed-2.0-pro}. In addition, to ensure reproducibility, we added the open-source model Gemma 4-31B \citep{Gemma4} as a judge model to score the generated results. 
The judge model reads the initial state image, the instruction, and the model-generated target state image. It assigns an integer score from 1 to 5 based on the action execution (following the instruction), scene consistency, and physical plausibility, along with the reasoning for the score. A higher score indicates that the generated image closely matches the target required by the instruction. Then, the average of all the scores will be calculated to obtain the percentage, which is the final score. The prompt used for benchmark evaluation is shown in Listing \ref{lst:evaluator_prompt}. 

\begin{center}
\begin{minipage}{0.90\linewidth}
\begin{lstlisting}[style=promptstyle, caption={Prompt used for benchmark evaluation.}, label={lst:evaluator_prompt}]
"You are a practical benchmark evaluator using lenient pass criteria. You will receive two images (an original image and a modified image) along with a specific modification instruction.

Modification instruction:
{instruction}

The first image is the original image. The second image is the modified image.

Score Instruction: Following from 1 to 5 (integers only). Score higher when the instruction's intended outcome is clearly visible. For agent-action instructions, the outcome should look executed--not merely teleported: if a visible person or robot should act, penalize cases where the result appears but the agent's pose, position, or contact state is essentially unchanged. Allow in-progress or imperfect execution. Minor occlusion or detail loss is fine.

General scoring philosophy:
- Use the full 1-5 range to reflect how well the instruction is followed.
- Minor blur, texture shifts, small artifacts, or partial ambiguity should NOT automatically fail.

Respond with JSON only, no markdown fences:
{
  "score": 3,
  "reasoning": "..."
}"
\end{lstlisting}
\end{minipage}
\end{center}

Specifically, the focus is on the following three aspects. First, the generated image should be in the same scene as the input image, preserving the original environment, viewpoint, and layout of major objects as much as possible. Second, the generated image should accurately reflect the state changes corresponding to the action command, such as the result of an object being moved, opened, picked up, or lifted. Third, the generated image should not contain content that obviously violates the physical laws, such as generating irrelevant objects out of thin air or lacking a reasonable causal relationship in the interaction process. If the state of an object changes, but the posture, position, or contact relationship of the executing subject remains almost unchanged, points will be deducted appropriately. Slight blurring, texture changes, small-scale artifacts, or local ambiguity will not be considered failures.

\subsubsection{Baselines} \label{sec:appendix_evaluation_toV_baselines}
In terms of image prediction, we use the following baselines for evaluation:

\vspace{-1ex}
\begin{itemize}
    \item \textbf{OmniGen2} \citep{OmniGen2} is a versatile open-source generative model for unified image generation, editing, and in-context generation. It introduces separate readout pathways for text and image modalities, a decoupled image tokenizer, and task-specific data construction pipelines for image editing and in-context generation.
    \item \textbf{FLUX.1-Kontext} \citep{FLUX.1-Kontext} is a generative flow-matching model for in-context image generation and editing. It takes both text and image inputs as context and supports local editing, global editing, character reference, style reference, and text editing within a unified architecture. 
    \item \textbf{FLUX.2 [klein]} \citep{flux2} is an efficient image generation and editing model. It adopts a rectified-flow Transformer architecture and supports text-conditioned generation as well as multi-reference image editing. We include it as a generative baseline to assess the visual synthesis and editing capability of compact world-oriented image models.
\end{itemize}

\subsection{Action Generation}
\label{sec:appendix_evaluation_toA}

\subsubsection{Real-Robot Benchmark}
\label{sec:appendix_evaluation_toA_Benchmark}

\begin{figure}[!htbp]
    \centering
    \includegraphics[width=0.98\linewidth]{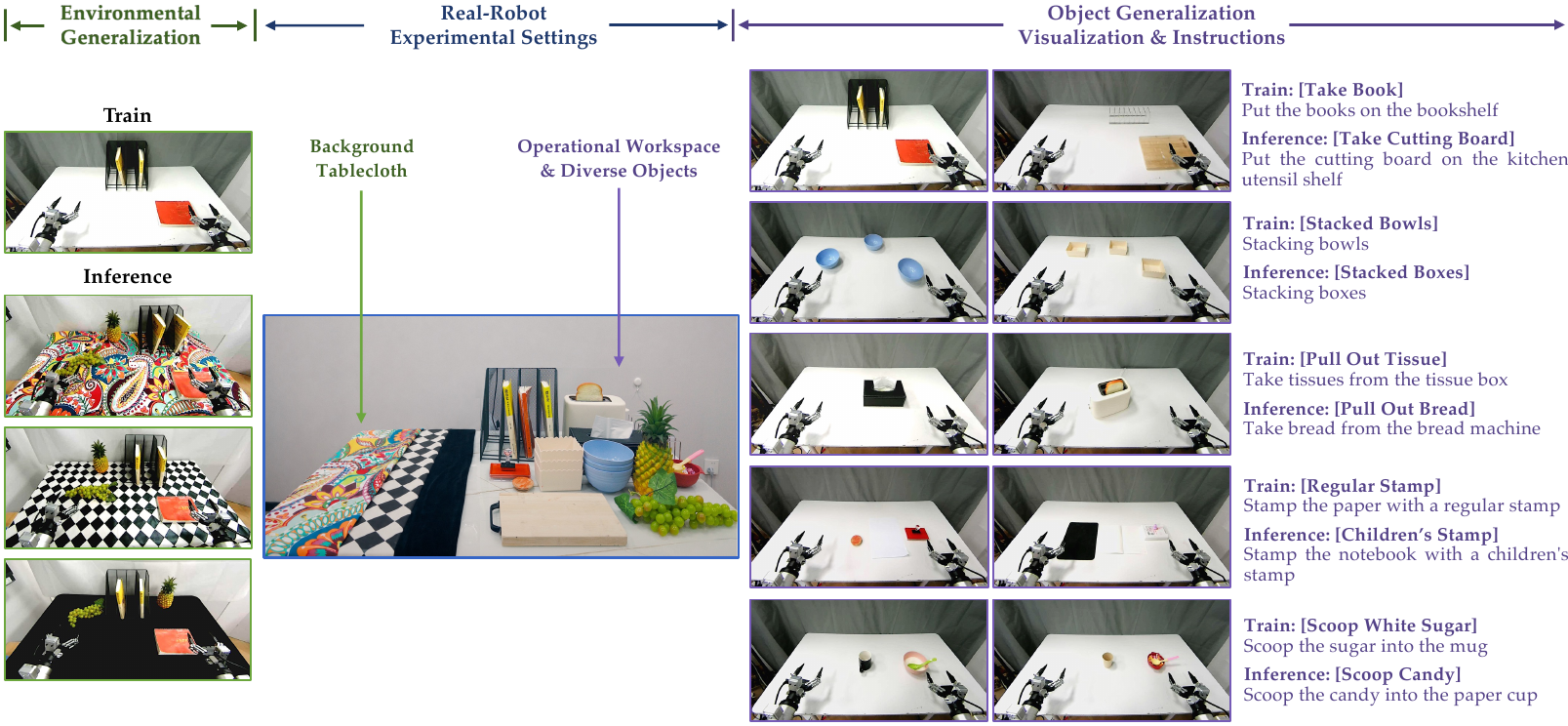}
    \caption{
    \textbf{Real-robot benchmark.}
    We evaluate the dual-arm wheeled robot on five manipulation tasks and construct OOD settings for environment and object generalization.
    }
    \label{fig:appendix_realrobot}
\end{figure}

We evaluate action readout on the dual-arm wheeled humanoid real robot with five manipulation tasks: \textit{Take Book}, \textit{Stacked Bowls}, \textit{Pull Out Tissue}, \textit{Stamp}, and \textit{Scoop Sugar}. 
For each task, we collect 200 real-robot trajectories for downstream \textit{Action Expert} post-training. 
The benchmark settings are shown in Figure~\ref{fig:appendix_realrobot}.

We construct two types of real-robot OOD settings. 
For \textit{environment OOD}, we keep the task objects and instructions unchanged, but vary the scene appearance using three unseen tablecloth/background settings. 
For \textit{object OOD}, we replace the task objects or target containers with unseen but semantically related instances. 
The object OOD settings are summarized in Table~\ref{tab:appendix_object_ood_settings}.

\begin{table}[!htbp]
  \centering
  \caption{\textbf{Object OOD settings for real-robot evaluation.}}
  \label{tab:appendix_object_ood_settings}
  \small
  \setlength{\tabcolsep}{4pt}
  \renewcommand{\arraystretch}{1.08}
  \begin{tabularx}{\linewidth}{
    @{}
    >{\raggedright\arraybackslash}p{0.13\linewidth}
    >{\raggedright\arraybackslash}p{0.35\linewidth}
    >{\raggedright\arraybackslash}X
    @{}
  }
    \toprule[1pt]
    \textbf{Task} & \textbf{Training Instruction} & \textbf{Inference Instruction} \\
    \midrule[0.5pt]
    \textit{Take Book} 
    & Put the book on the bookshelf. 
    & Put the cutting board on the kitchen utensil shelf. \\

    \textit{Stacked Bowls} 
    & Stack bowls. 
    & Stack boxes. \\

    \textit{Pull Out Tissue} 
    & Take tissue from the tissue box. 
    & Take bread from the bread machine. \\

    \textit{Stamp} 
    & Stamp the paper with a regular stamp. 
    & Stamp the notebook with a children's stamp. \\

    \textit{Scoop Sugar} 
    & Scoop sugar into the mug with a spoon. 
    & Scoop candy into the paper cup with a spoon. \\
    \bottomrule[1pt]
  \end{tabularx}
\end{table}

\subsubsection{Metrics}
\label{sec:appendix_evaluation_toA_metrics}
We evaluate real-robot performance from two complementary perspectives. First, we use task-specific rule-based scores. As shown in Table~\ref{tab:real_robot_scoring}, each task is decomposed into several key stages, and the rule-based score measures the highest completed stage before termination. Second, we adopt PRM-as-a-Judge~\citep{PRM-as-a-Judge} to provide dense trajectory-level diagnostics of progress and execution quality.

\begin{table}[!ht]
\centering
\caption{
\textbf{Scoring criteria for real-robot evaluation.}
Each task is evaluated within 60 seconds. 
If the robot becomes locked due to severe collision, or if the object falls and the task can no longer continue, evaluation is stopped. 
For each task, only the highest achieved score before termination is counted.
}
\label{tab:real_robot_scoring}
\small
\renewcommand{\arraystretch}{0.83}
\begin{tabularx}{\textwidth}{C{1.3cm}Y C{0.6cm}}
\toprule
Task & \multicolumn{1}{c}{Scoring Criteria} & Point \\
\midrule

\multirow{6}{1.5cm}{\centering Take Book}
& 1. The robot arm moves toward the book. & 10 \\
& 2. The gripper contacts the book. & 10 \\
& 3. The book is pushed to the edge, with more than 2 cm beyond the edge, without falling. & 20 \\
& 4. The book is successfully grasped. & 30 \\
& 5. The book is moved toward the bookshelf while being grasped. & 20 \\
& 6. The book is successfully placed on the bookshelf. & 10 \\
\midrule

\multirow{9}{1.5cm}{\centering Stacked Bowls}
& 1. The hand moves toward Bowl 1. & 10 \\
& 2. Bowl 1 is grasped. & 20 \\
& 3. Bowl 1 is placed stably. & 10 \\
& 4. The hand moves toward Bowl 2. & 10 \\
& 5. Bowl 2 is grasped. & 10 \\
& 6. Bowl 2 is stably stacked into Bowl 1. & 10 \\
& 7. The hand moves toward Bowl 3. & 10 \\
& 8. Bowl 3 is grasped. & 10 \\
& 9. Bowl 3 is stably stacked into Bowl 2. & 10 \\
\midrule

\multirow{6}{1.5cm}{\centering Pull Out Tissue}
& 1. Arm A moves toward the tissue box. & 10 \\
& 2. Arm A holds the tissue box. & 20 \\
& 3. Arm B moves toward the tissue. & 20 \\
& 4. Arm B successfully grasps the yellow tissue. & 40 \\
& 5. The tissue is placed on the table. & 10 \\
& $\triangleright$ \textbf{The two arms are scored separately.} & - \\
\midrule

\multirow{7}{1.5cm}{\centering Stamp}
& 1. The robot arm moves toward the stamp. & 10 \\
& 2. The stamp is successfully grasped and lifted. & 30 \\
& 3. The stamp is moved above the document. & 10 \\
& 4. The document is stamped by pressing the stamp. & 20 \\
& 5. The stamp is moved above the ink pad. & 10 \\
& 6. The stamp is placed stably without toppling. & 20 \\
& $\triangleright$ \textbf{If the stamp topples, scoring stops.} & - \\
\midrule

\multirow{6}{1.5cm}{\centering Scoop Sugar}
& 1. The hand moves toward the spoon. & 10 \\
& 2. The spoon is successfully grasped. & 20 \\
& 3. Sugar is scooped with the spoon. & 20 \\
& 4. The spoon is moved to the mug; the spoon must be held, but sugar is not strictly required. & 10 \\
& 5. The sugar is poured into the mug; the spoon must be held, but sugar is not required. & 20 \\
& 6. The spoon is placed back on the right side of the table. & 20 \\

\bottomrule
\end{tabularx}
\end{table}

\subsubsection{Baselines}
\label{sec:appendix_evaluation_toA_baselines}

We compare Orca with the following baselines in real-robot embodied tasks:

\vspace{-2ex}
\begin{itemize}
    \item \textbf{V-JEPA 2.1}~\citep{V-JEPA-2.1}. 
    The native V-JEPA-AC planner requires a goal image, which is unavailable in our real-robot OOD setting. 
    Therefore, we use the hidden representation of V-JEPA 2.1 as the condition for the same downstream action expert used by Orca.

    \item \textbf{Qwen3.5}~\citep{qwen35}. 
    To clarify performance attribution, we use the last-layer hidden state of Qwen3.5 as the condition for the same downstream action expert. 
    This comparison tests whether Orca's learned world representation provides more useful information for action readout than a general vision-language representation.

    \item \textbf{$\pi_{0.5}$}~\citep{pi0.5}. 
    We use $\pi_{0.5}$ as a strong VLA baseline pretrained on large-scale robot data. 
    This comparison evaluates whether Orca's learned world representation can provide competitive or complementary benefits under limited real-robot trajectories.
\end{itemize}

During post-training, the V-JEPA 2.1 backbone, the Qwen3.5 backbone, and the VLM component of $\pi_{0.5}$ are frozen, and only the action expert is trained. 
For V-JEPA 2.1, Qwen3.5, and Orca, the action experts are configured identically, initialized from scratch, and trained on 200 trajectories per task. 
We train each task for 20k steps with a global batch size of 128, which gives the best empirical performance among the settings we tested. 
For $\pi_{0.5}$, we follow its official post-training configuration and train for 30k steps with a global batch size of 32; we also find this setting to yield the best performance for $\pi_{0.5}$ in our benchmark.

\subsubsection{Detailed Real-Robot Results}
\label{sec:appendix_evaluation_toA_results}
The PRM-as-a-Judge results are used in the main-text analysis, while this appendix provides the detailed task-level rule-based evaluation protocol and results. 
Table~\ref{tab:appendix_detailed_real_robot_ood} provides the OOD results. 
The rule-based scores report task-level completion under the manually designed scoring criteria.

\begin{table*}[!htbp]
  \centering
  \vspace{-0.5em}
  \caption{\textbf{Detailed rule-based results under real-robot OOD settings.}}
  \label{tab:appendix_detailed_real_robot_ood}
  \setlength{\tabcolsep}{4pt}
  \renewcommand{\arraystretch}{1.05}
    \begin{tabularx}{\textwidth}{
      >{\centering\arraybackslash}p{0.22\textwidth}|
      >{\raggedright\arraybackslash}p{0.17\textwidth}|
      *{6}{>{\centering\arraybackslash}X}
    }
    \toprule[1pt]
    \multirow{2}{*}{Settings} 
    & \multirow{2}{*}{Model}
    & \multicolumn{6}{c}{Rule-based Score} \\
    \cmidrule(lr){3-8}
    & 
    & Book & Bowls & Tissue & Stamp & Sugar & Average \\
    \midrule[0.5pt]

    \multirow{6}{*}{Environment OOD}
    & $\pi_{0.5}$
    & 27 & 44 & 32 & 9 & 26 & 27.6\\
    & V-JEPA 2.1
    & 24 & 15 & 28 & 6 & 3 & 15.2\\
    & Qwen3.5-0.8B
    & 1 & 28 & 0 & 0 & 10 & 7.8\\
    & Qwen3.5-4B
    & 19 & 27 & 0 & 6 & 10 & 12.4\\

    \rowcolor[rgb]{.89,.949,.851}
    & \textbf{Orca-0.8B}
    & 23 & 44 & 28 & 27 & 15 & 27.4\\
    \rowcolor[rgb]{.89,.949,.851}
    & \textbf{Orca-4B}
    & 25 & 41 & 39 & 62 & 16 & 36.6\\

    \midrule[0.5pt]

    \multirow{6}{*}{Object OOD}
    & $\pi_{0.5}$
    & 30 & 46 & 55 & 13 & 12 & 31.2\\
    & V-JEPA 2.1 
    & 34 & 0 & 44 & 2 & 14 & 18.8\\
    & Qwen3.5-0.8B
    & 3 & 27 & 0 & 0 & 10 & 8.0\\
    & Qwen3.5-4B
    & 6 & 23 & 0 & 0 & 14 & 8.6\\

    \rowcolor[rgb]{.89,.949,.851}
    & \textbf{Orca-0.8B}
    & 24 & 23 & 65 & 12 & 10 & 26.8\\
    \rowcolor[rgb]{.89,.949,.851}
    & \textbf{Orca-4B}
    & 34 & 28 & 59 & 4 & 16 & 28.2\\

    \bottomrule[1pt]
  \end{tabularx}
\end{table*}

\subsubsection{Additional Qualitative Visualizations}
\label{sec:appendix_evaluation_toA_visualization}

\begin{figure}[!htbp]
    \centering
    \includegraphics[width=1\linewidth]{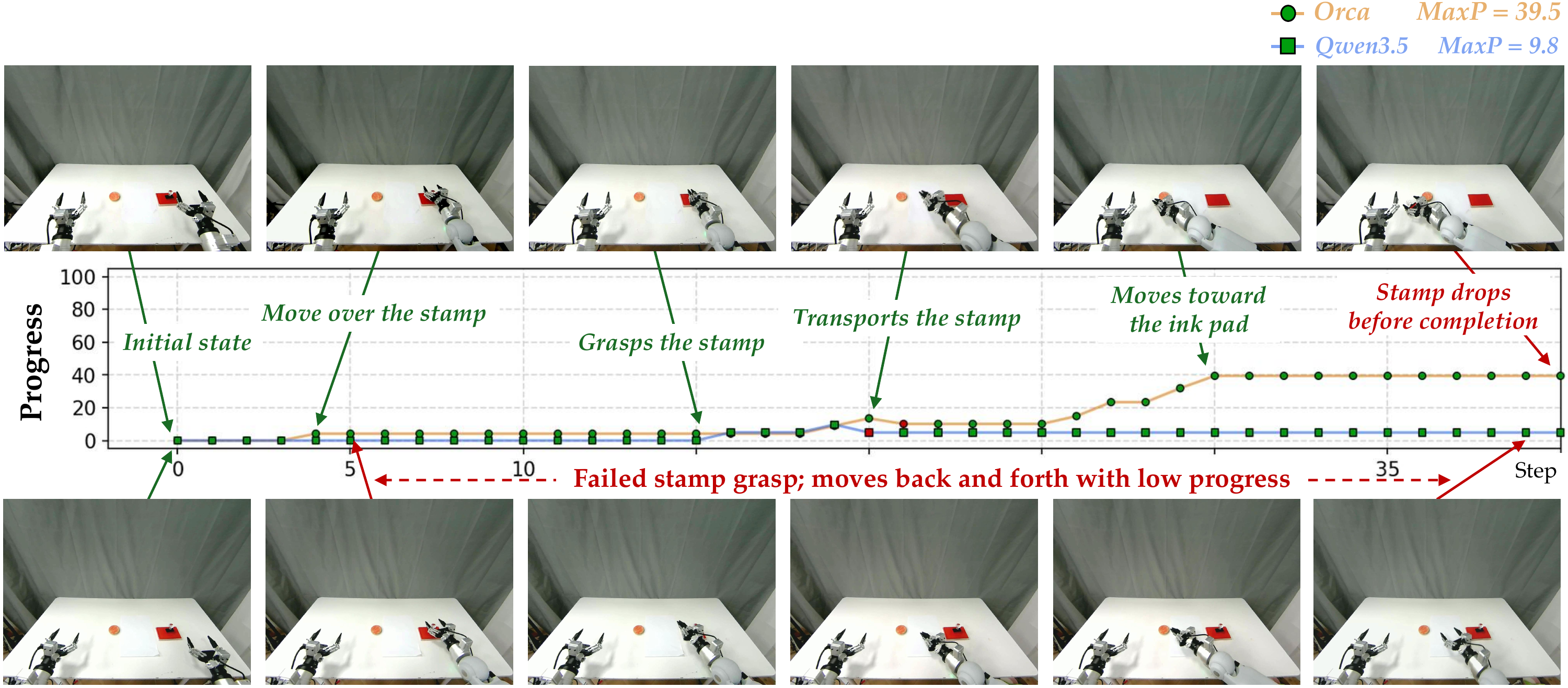}
\caption{
\textbf{Failure with higher intermediate progress in Stamp.}
Orca grasps and transports the stamp toward the ink pad before dropping it near the end, while Qwen3.5 fails to maintain a meaningful stamp grasp and remains at low progress.
}
    \label{fig:toa_con1_case1}
\end{figure}

\begin{figure}[!htbp]
    \centering
    \includegraphics[width=1\linewidth]{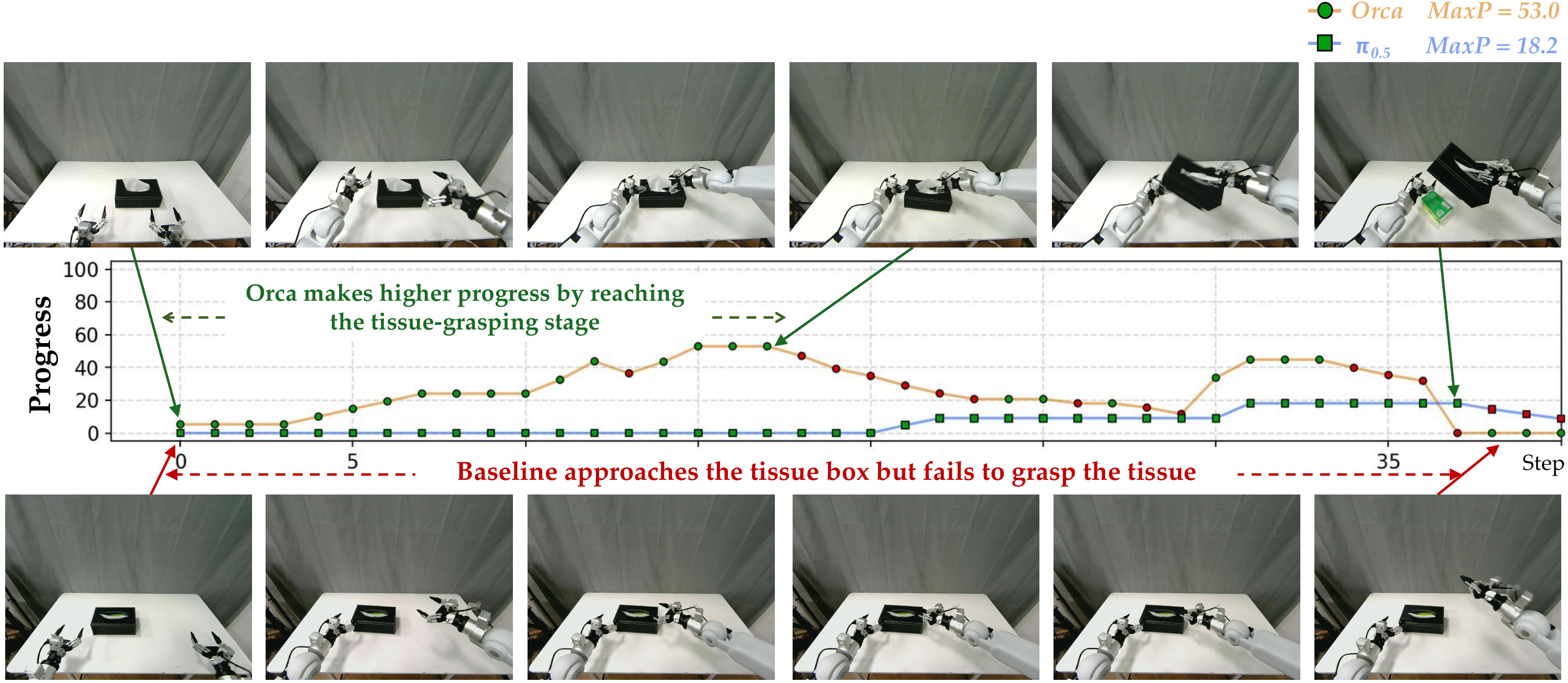}
\caption{
\textbf{Failure with higher intermediate progress in Pull Out Tissue.}
Orca reaches the tissue-grasping stage and achieves substantially higher intermediate progress, while $\pi_{0.5}$ only approaches the tissue box and fails to grasp the tissue.
}
\label{fig:toa_con1_case2}
\end{figure}

\begin{figure}[!htbp]
    \centering
    \includegraphics[width=0.95\linewidth]{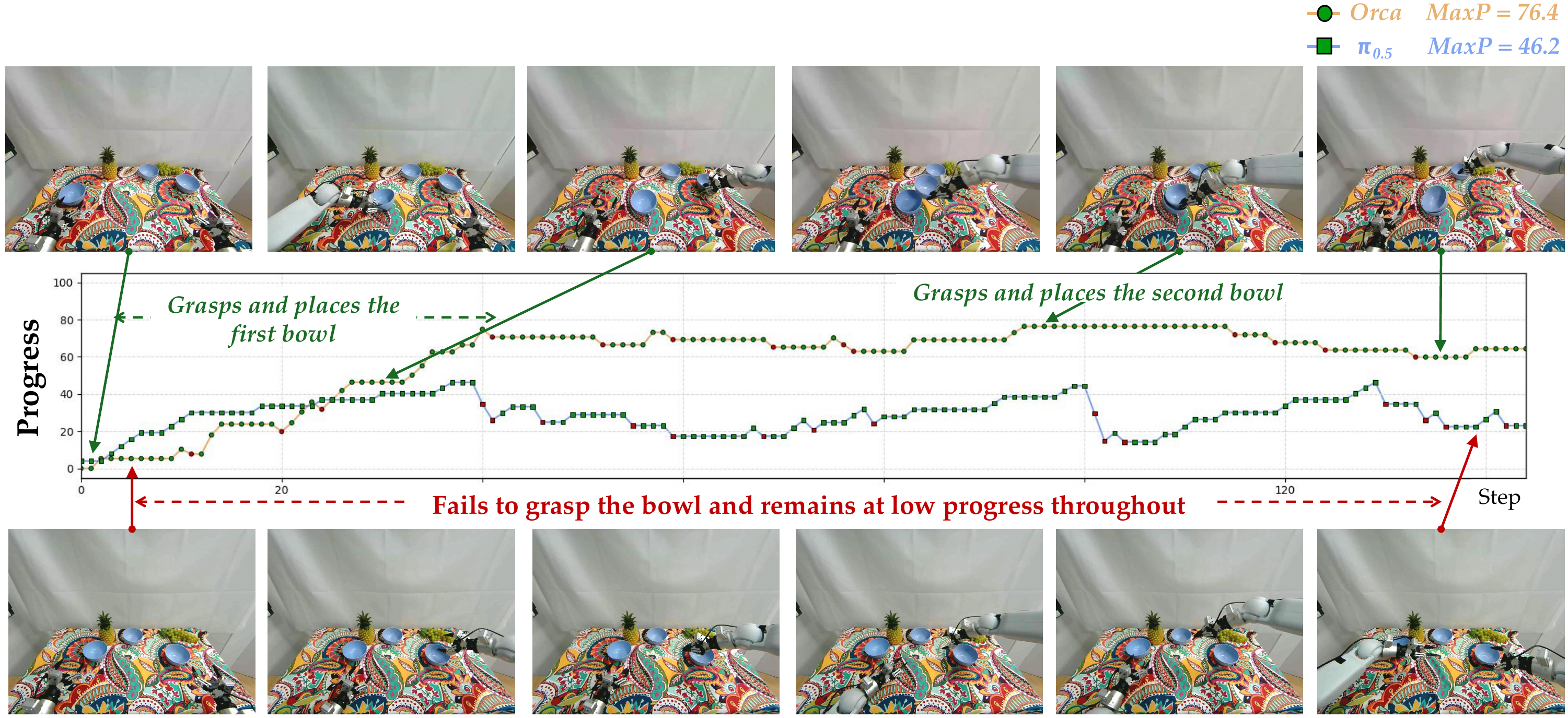}
    \vspace{-0.8em}
\caption{
\textbf{Failure with higher intermediate progress in Stacked Bowls.}
Orca advances through multiple bowl-stacking stages, while $\pi_{0.5}$ repeatedly fails to grasp the bowl and remains at lower progress.
}
    \label{fig:toa_con2_case1}
    \vspace{-1em}
\end{figure}

\begin{figure}[!htbp]
    \centering
    \includegraphics[width=0.95\linewidth]{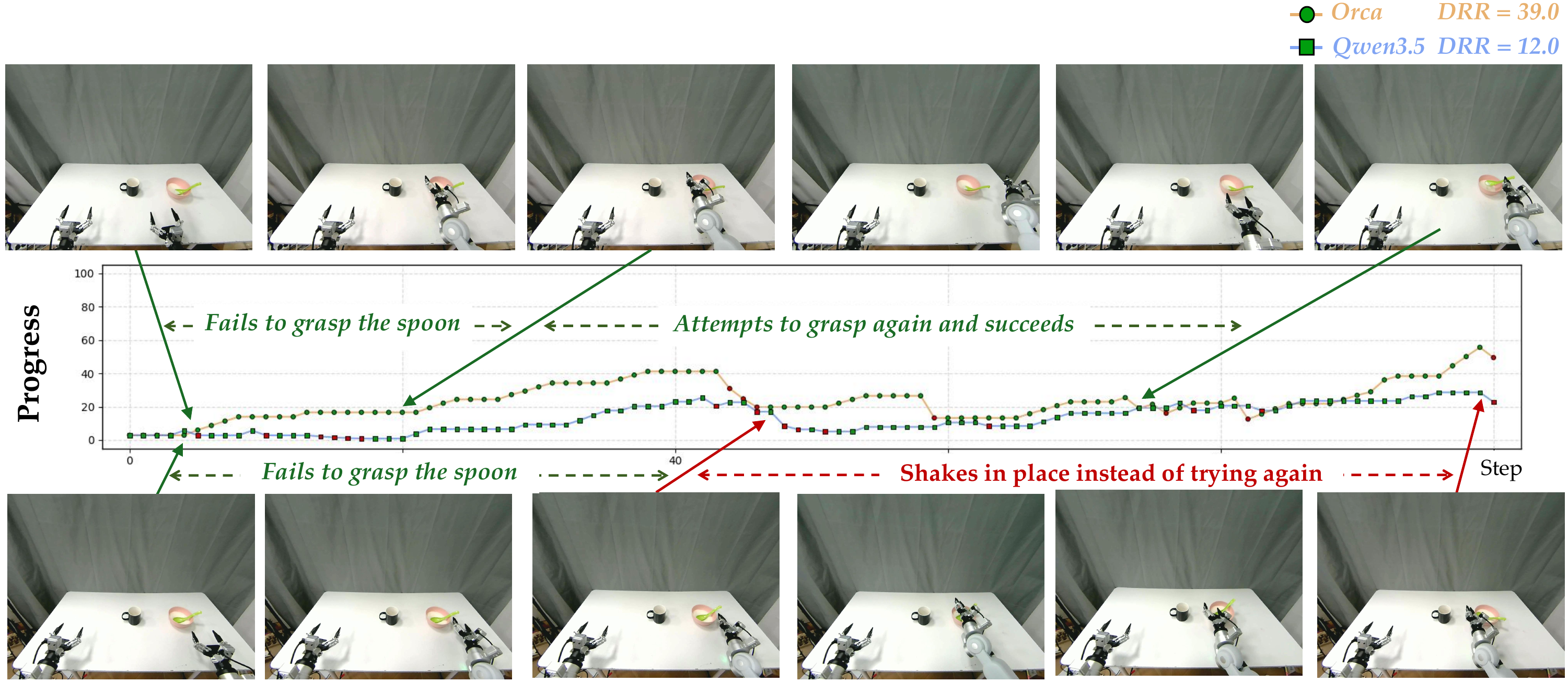}
    \vspace{-0.8em}
\caption{
\textbf{Partial recovery after spoon-grasp failure in Scoop Sugar.}
Orca retries after failing to grasp the spoon and recovers some lost progress, while Qwen3.5 shakes in place without effective re-grasping.
}
    \label{fig:toa_con2_case2}
    \vspace{-1em}
\end{figure}

\begin{figure}[!htbp]
    \centering
    \includegraphics[width=0.95\linewidth]{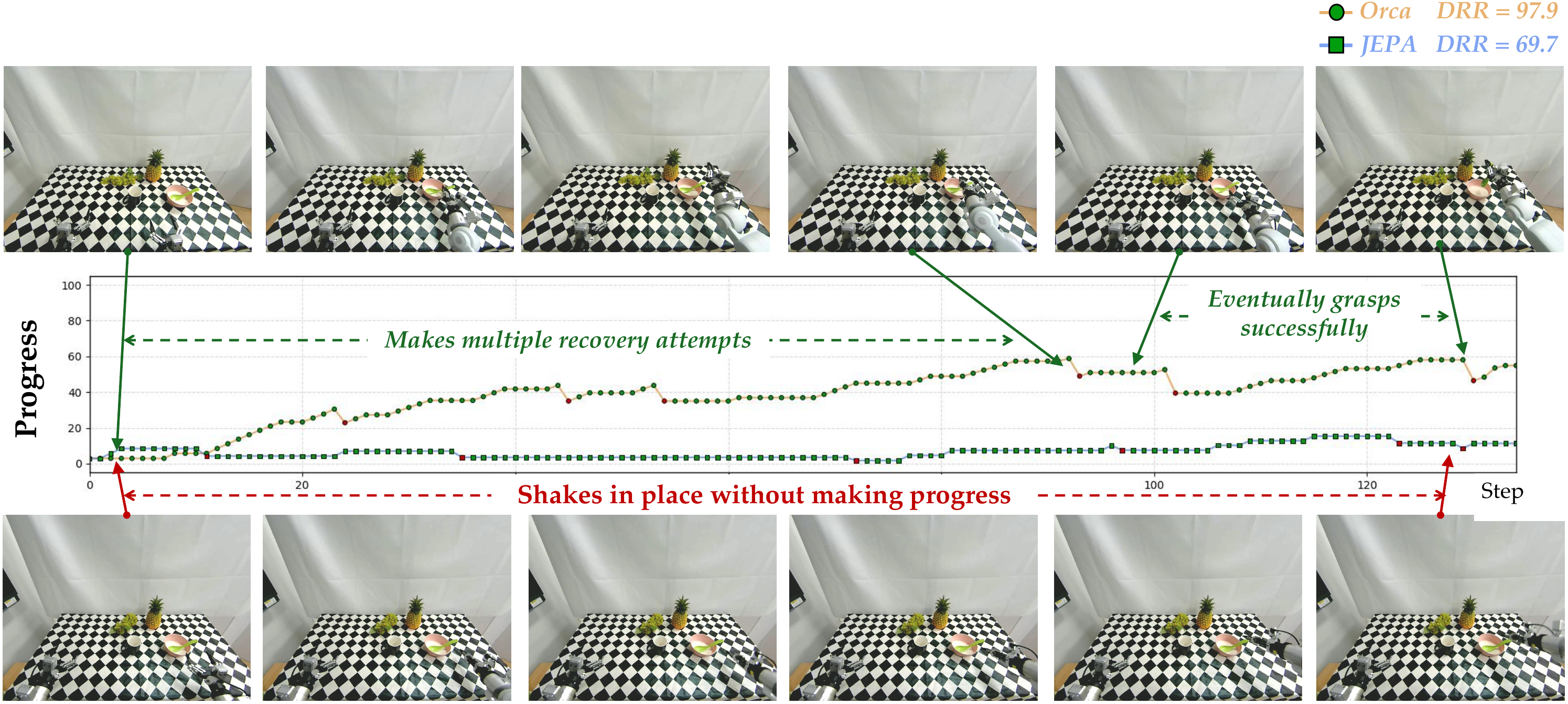}
    \vspace{-0.8em}
\caption{
\textbf{Recovery through repeated spoon-grasp attempts in Scoop Sugar.}
Orca makes multiple recovery attempts and eventually grasps the spoon successfully, while JEPA remains largely stagnant with limited task progress.
}
    \label{fig:toa_con2_case3}
    \vspace{-2em}
\end{figure}

In addition to the qualitative example shown in the main text, we provide more trajectory visualizations in Figure~\ref{fig:toa_con1_case1}-Figure~\ref{fig:toa_con2_case3}. 
These examples further illustrate two process-level advantages of Orca: maintaining higher progress even in failed trajectories and recovering from intermediate grasp failures.

\clearpage

\section{More Visualization}\label{sec:appendix_visualization}
\setnavsection{sec:appendix_visualization}

\renewcommand{\thefigure}{F\arabic{figure}}
\renewcommand{\theHfigure}{F\arabic{figure}}
\setcounter{figure}{0}

\renewcommand{\thetable}{F\arabic{table}}
\renewcommand{\theHtable}{F\arabic{table}}
\setcounter{table}{0}

\subsection{Cross-Benchmark Capability Analysis for Text Generation}\label{sec:appendix_benchmark}

As shown in \textbf{Section \ref{sec:evaluation_toL}}, we identify a set of generalized, high-level capability dimensions that transcend benchmark boundaries, namely state transition, commonsense reasoning, spatial relations, and dynamic motion.
Several representative examples are provided below.

\begin{figure}[!htbp]
    \centering
    \includegraphics[width=0.94\linewidth]{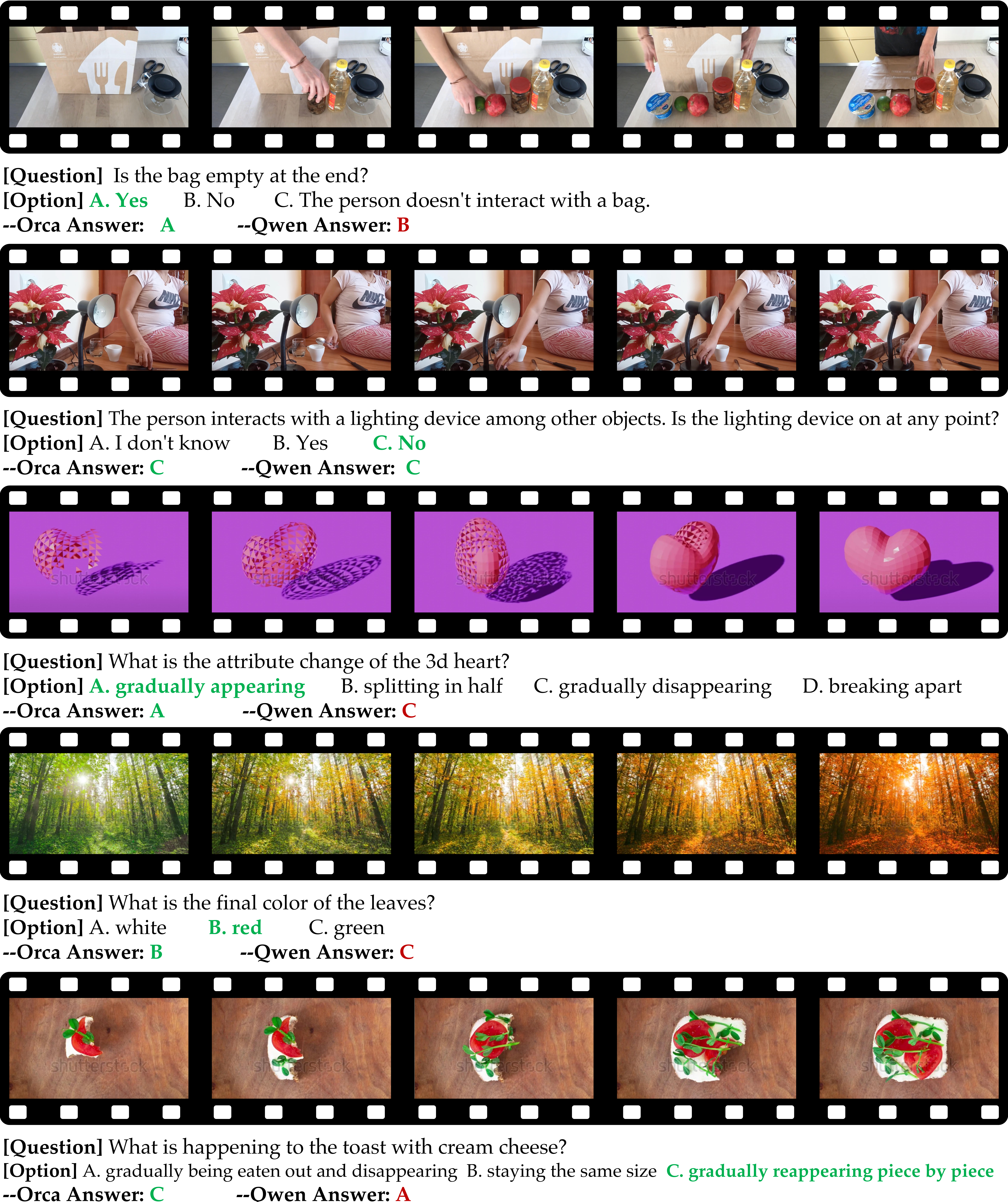}
    \caption{\textbf{Cross-benchmark examples of state transition.} This dimension evaluates a model’s understanding of causal temporal dynamics and physical state changes, namely its ability to predict or recognize the evolution of an object from state A to state B. The improvement is particularly evident in tasks involving irreversible physical processes.}
    \label{fig:state-cross-bench1}
\end{figure}

\begin{figure}[!htbp]
    \centering
    \includegraphics[width=0.75\linewidth]{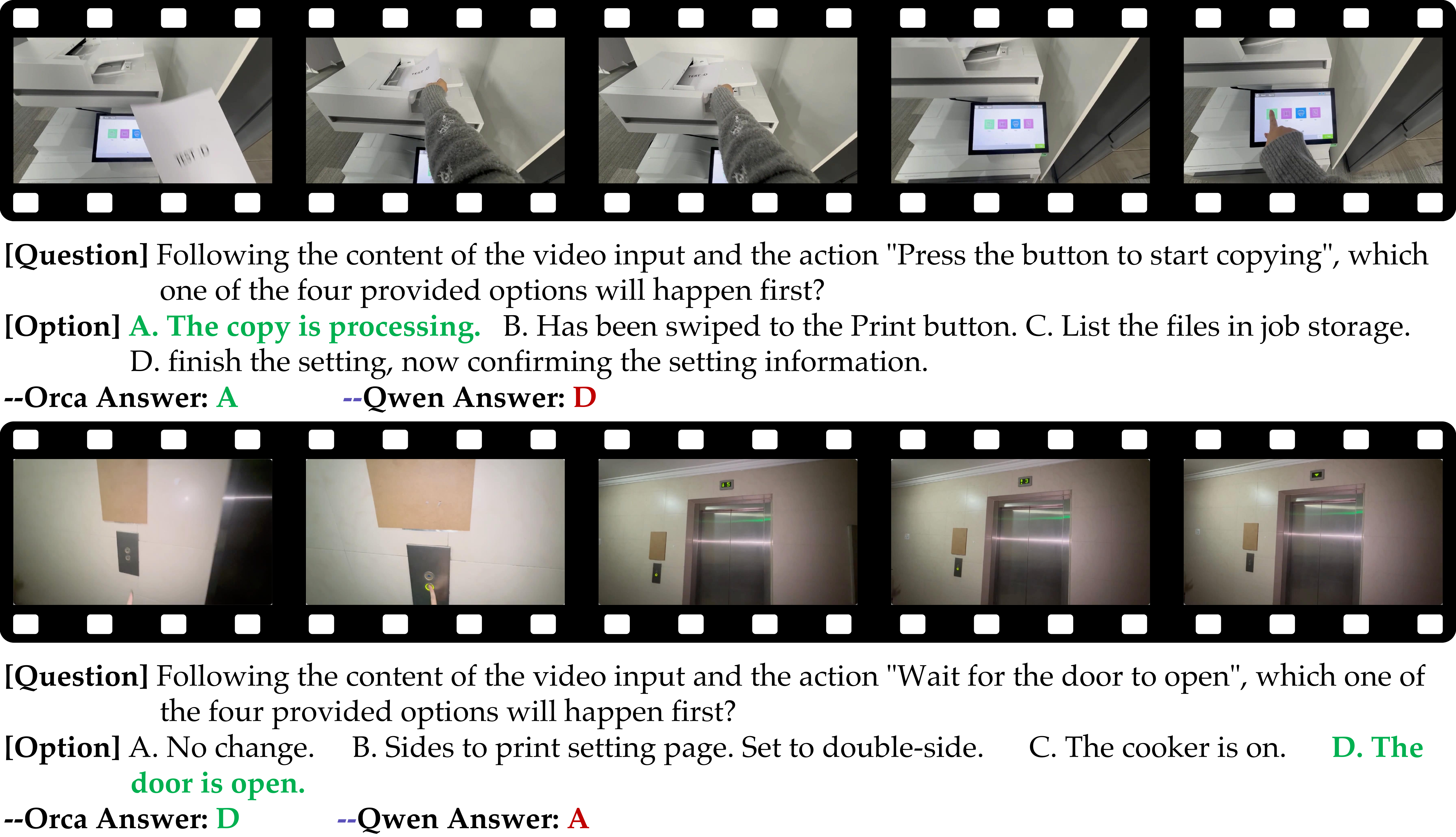}
  \caption{\textbf{More cross-benchmark examples of state transition.}}
  \label{fig:state-cross-bench2}
\end{figure}

\begin{figure}[!htbp]
    \centering
    \includegraphics[width=0.75\linewidth]{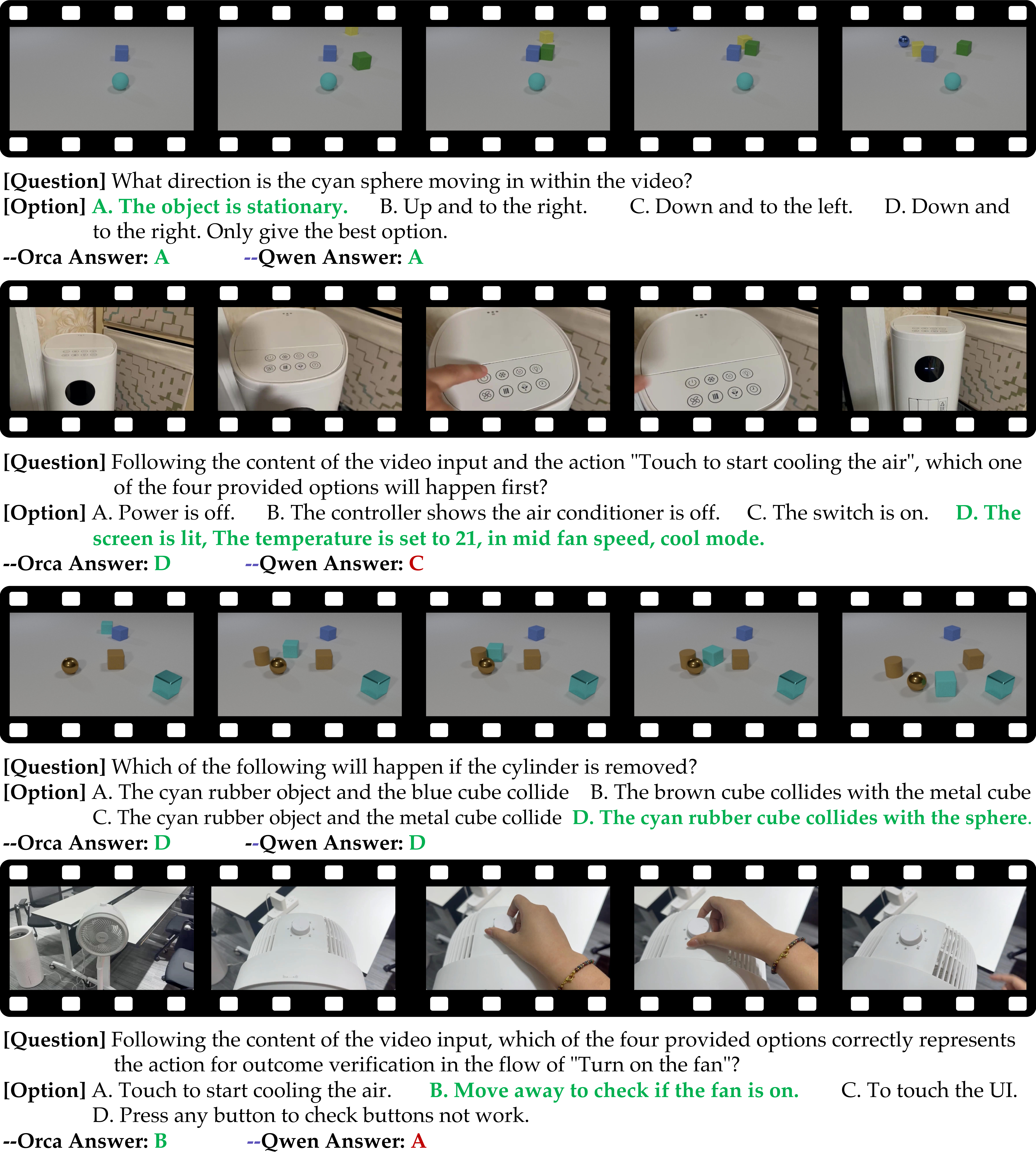}
    \caption{\textbf{Cross-benchmark examples of commonsense reasoning.} The advantage of Orca is particularly pronounced in complex VQA scenarios that require reasoning beyond the visible scene and inferring hypothetical outcomes.}
    \label{fig:sense-cross-bench}
\end{figure}

\begin{figure}[!htbp]
    \centering
    \includegraphics[width=0.95\linewidth]{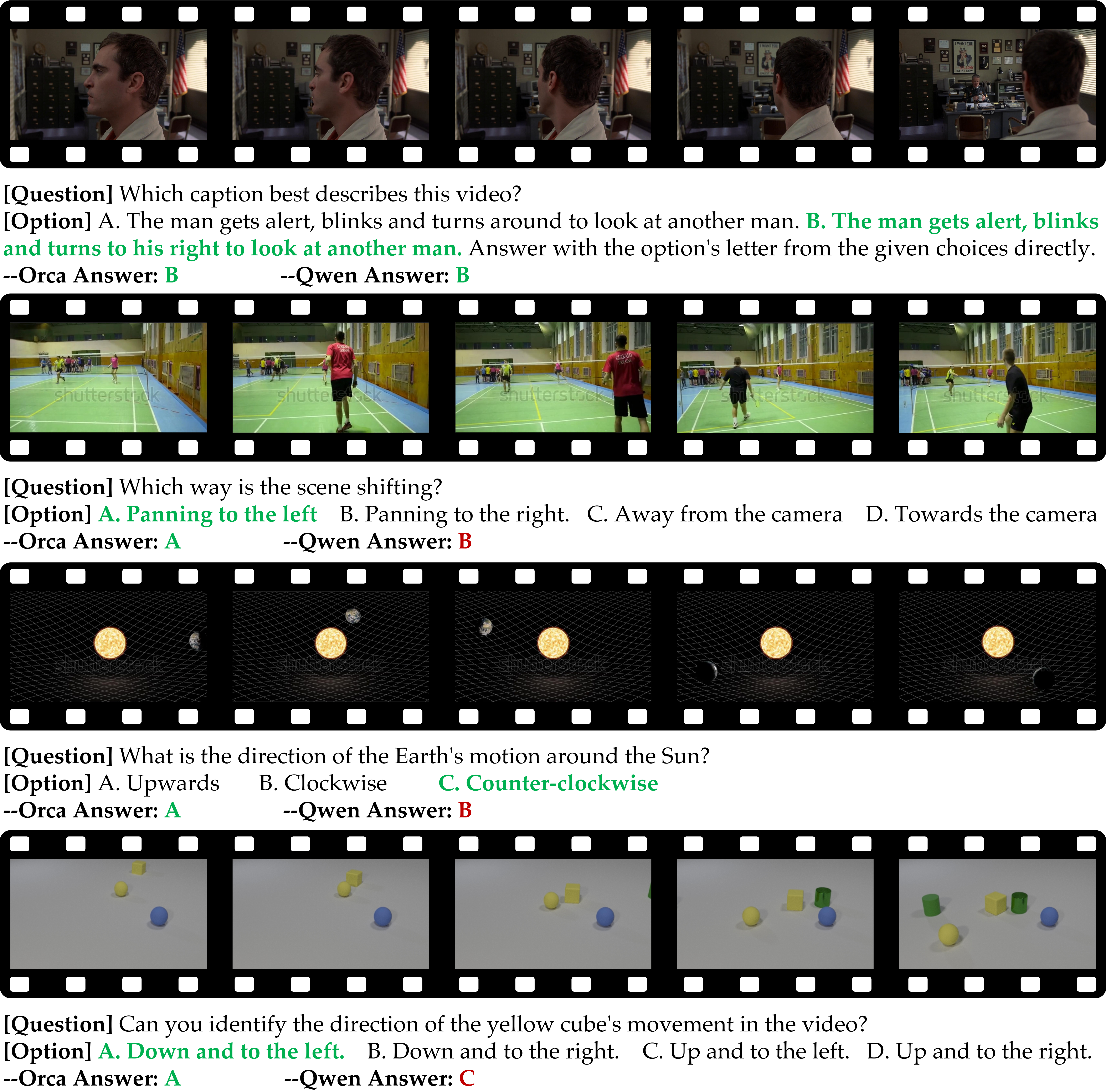}
    \caption{\textbf{Cross-benchmark examples of dynamic motion.} The proposed unconscious learning paradigm enables Orca to naturally acquire temporal continuity and motion inertia, leading to stronger forward simulation capabilities for dynamic object behaviors.}
    \label{fig:move-cross-bench}
\end{figure}

\begin{figure}[!htbp]
    \centering
    \includegraphics[width=1\linewidth]{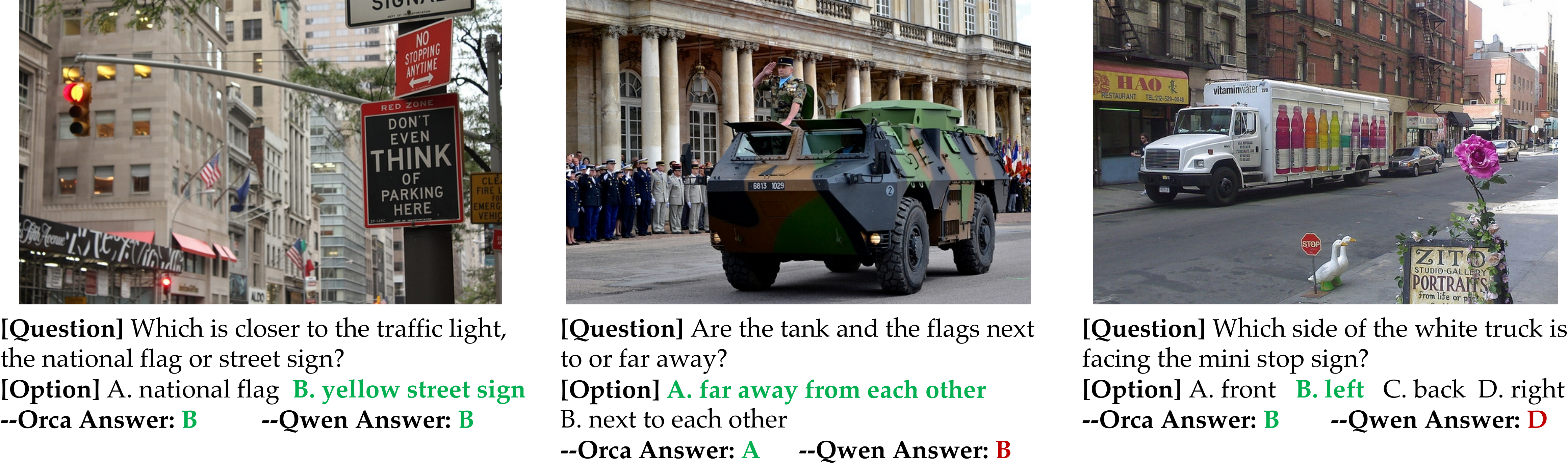}
    \caption{\textbf{Cross-benchmark examples of spatial relations.} The results demonstrate strong robustness in scenarios involving complex occlusions and multi-object spatial reasoning.}
    \label{fig:spatial-cross-bench}
\end{figure}

\end{document}